\documentclass{article}

\usepackage[utf8]{inputenc} % allow utf-8 input
\usepackage[T1]{fontenc}    % use 8-bit T1 fonts
\usepackage{hyperref}       % hyperlinks
\usepackage{url}            % simple URL typesetting
\usepackage{booktabs}       % professional-quality tables
\usepackage{amsfonts}       % blackboard math symbols
\usepackage{nicefrac}       % compact symbols for 1/2, etc.
\usepackage{microtype}      % microtypography
\usepackage{xcolor}         % colors
\usepackage{subfigure}
\usepackage{amsmath,amsfonts}
\usepackage{algorithmic}
\usepackage{array}
\usepackage[margin=0.5in]{geometry}
\usepackage{textcomp}
\usepackage{stfloats}
\usepackage{url}
\usepackage{verbatim}
\usepackage{graphicx}
\usepackage{xcolor}
\usepackage{float}
\usepackage{authblk}
\def\BibTeX{{\rm B\kern-.05em{\sc i\kern-.025em b}\kern-.08em
    T\kern-.1667em\lower.7ex\hbox{E}\kern-.125emX}}
\usepackage{balance}
\usepackage{multicol}

\title{ConvFormer: Parameter Reduction in Transformer Models for 3D Human Pose Estimation by Leveraging Dynamic Multi-Headed Convolutional Attention}
\author{
  Alec Diaz-Arias and Dmitriy Shin \\
  Inseer, Inc.\\
  Iowa City, IA 52241 \\
  alec.diaz-arias@inseer.com \\
}

\begin{document}

\maketitle

\abstract{    Recently, fully-transformer architectures have replaced the defacto convolutional architecture for the 3D human pose estimation task. In this paper we propose \textbf{\textit{ConvFormer}}, a novel convolutional transformer that leverages a new \textbf{\textit{dynamic multi-headed convolutional self-attention}} mechanism for monocular 3D human pose estimation. We designed a spatial and temporal convolutional transformer to comprehensively model human joint relations within individual frames and globally across the motion sequence. Moreover, we introduce a novel notion of \textbf{\textit{temporal joints profile}} for our temporal ConvFormer that fuses complete temporal information immediately for a local neighborhood of joint features. We have quantitatively and qualitatively validated our method on three common benchmark datasets: Human3.6M, MPI-INF-3DHP, and HumanEva. Extensive experiments have been conducted to identify the optimal hyper-parameter set. These experiments demonstrated that we achieved a \textbf{significant parameter reduction relative to prior transformer models} while attaining State-of-the-Art (SOTA) or near SOTA on all three datasets. Additionally, we achieved SOTA for Protocol III on H36M for both GT and CPN detection inputs. Finally, we obtained SOTA on all three metrics for the MPI-INF-3DHP dataset and for all three subjects on HumanEva under Protocol II. }

\maketitle

\section{Introduction}
Monocular 3D \textit{Human Pose Estimation} (HPE) is a process of localizing joint locations  and subsequently a body representation (skeletal representation) from varying input streams such as a static image  or a video stream. 3D HPE receives a lot of attention in the computer vision community and plays an essential role in many applications including motion analysis, computer animation, action recognition, and ergonomic risk-safety assessment. Many approaches had been proposed to solve this problem (for an extended treatment see Prior Works section).

More recently, and motivated by the ground-breaking work in \cite{KDWHUBMDHGUZ21} (ViT) a fully transformer architecture was introduced for the 3D HPE task. Transformers were developed to exploit long-range dependencies and have achieved tremendous success in NLP since their invention \cite{VSPUJGKP17} and more recently in various CV tasks.

All of these methods have continued to push the boundaries for accuracy, however, they have continually done so by increasing the network capacity \cite{PA21,CJHR21,YCWYSJTFY21,HMMFA21} and potentially leading to over-parametrization. For instance, classic transformers suffer from a known problem of redundancy which occurs due to complete connectivity. There have been works in NLP that have sought to introduce sparsity and have seen substantial accuracy increases while reducing the computational complexity \cite{WLLLH20,kitaev2020reformer,Jaszczur2021SparseIE}. Given that classic transformers are still in their infancy for CV tasks, sparsity mechanisms have not been fully applied yet. Solving the redundancy problem of classic transformers was one of the main motivations to introduce ConvFormer. For the 3D HPE problem and specifically for human motion, individual joints may exhibit a high degree of inter-correlation. Thus, by generating queries, keys, and values via fully connected layers in vanilla transformers, the networks learn redundancies leading to noisier inference. ConvFormer leverages convolutions to extract combinations of joints via their local receptive field that together provide a stronger signal that is less susceptible to noise, results in fewer features for the attention computations, and extensively reduces parameter counts. Furthermore, a single filter may be incapable of fully capturing dependencies. For this reason we introduced a dynamic aggregation mechanism which weights the contribution of different joint neighborhoods. We call this novel mechanism \textbf{\textit{dynamic multi-headed convolutional self-attention}} (DMHCSA).

Following \cite{ZZMYCD21, li2022mhformer} we leverage a spatial-temporal framework. However, a critical distinguishing factor and substantial motivation for ConvFormer was how to extract temporal dependencies at the query, key, value level prior to computing temporal attention maps. To achieve this we introduce \textbf{\textit{temporal joints profile}}. To compute them, ConvFormer extracts correlations between joints for individual frames in the spatial ConvFormer and subsequently generates high-order temporal profiles of joints present in the motion sequences with the temporal ConvFormer. More specifically for the temporal blocks, the DMHCSA mechanism extracts queries, keys, and values that have visibility across the motion sequence, which has been coined \textbf{\textit{early temporal fusion}}, leading to more complex self-attention maps that capture more intricate correlations. We conducted extensive experiments on three standard 3D HPE datasets, i.e.,  Human3.6M, MPI-INF-3DHP, and HumanEva-I \cite{IPOS13, mono-3dhp2017, SBB10} and compare ConvFormer against several competitive 3D HPE solution methodologies. ConvFormer achieves state-of-the-art results by the majority of the metrics and comparable on CPN detections under Protocol I while reducing the parameter count by more than half of models that take input sequences of equivalent length. Our contributions in this paper are summarized as follows:

\begin{enumerate}
    \item A significant parameter reduction relative to other transformer models using a new architecture called ConvFormer. ConvFormer leverages a novel multi-headed convolutional self-attention mechanism that dynamically aggregates sub-queries, keys, and values into a richer set of cues for 3D HPE.
    \item A novel notion of temporal joints profile is introduced that relies on immediate fusion of complete temporal information of the motion sequence.
    \item An extensive study on factors affecting the performance of ConvFormer.
\end{enumerate}

\section{Prior Works}
At the outset of leveraging deep neural networks for 3D HPE, many methods attempted to learn mappings from monocular RGB images to 3D skeletal representations \cite{PZDD17,SXWLW17}. While one-shot 3D HPE saw some success, it suffered from substantial computational overhead and simultaneously poor generalizability due to Motion Capture data being acquired in staged environments. In part due to Martinez et al. \cite{MHRL17}, the 3D HPE landscape shifted its focus predominantly towards the two-stage approach by leveraging accurate performance of off-the-shelf 2D pose detectors and then building networks that perform the 2D-to-3D lifting. A number of other works improved the performance of 3D HPE from a single monocular image utilizing various deep learning techniques and analytical approaches (e.g., \cite{SXWLW17,zhou2019hemlets,WHXYS20, D-AMSB21,ZHJJL21}).

To reduce errors, improve handling of self-occlusions, and increase generalizability of 3D HPE models, several works exploited spatial relationships among joints. To account for these relationships, some methods incorporate “static” anthropometric constraints and regularization procedures, while others are based on temporal architectures that infer these dependencies across video frames \cite{DMKASJ18,PFGA19,WYXL20, CFSZCL21}. For instance, people have used graph convolutional and graph attention networks to naturally model spatial relationships between joints while building lightweight networks \cite{VCCRLB18, SDMR20, WHXYS20, BGK21}.

Recently, Kolesnikov et al. introduced Vision Transformer (ViT) that applies a global self-attention mechanism to efficiently exploit salient information from video frames \cite{KDWHUBMDHGUZ21}. Since ViT, transformers have seen successes in many  CV tasks, including, image recognition \cite{KDWHUBMDHGUZ21,YCWYSJTFY21}, and object detection  \cite{CMSUKZ20}.

Even more recently, researchers have begun leveraging convolutions in transformers, to learn positional embeddings in place of dense layers, replacing the dense feed forward component with a convolutional feed-forward block for sparsity, or leveraging convolutional projections for specific tasks like video synthesis and working with unstructured point cloud data  \cite{ZSWJYCL20,CJHR21}. Finally, with the explosion of transformer-based models and their capability to effectively capture local and global relationships, they began to be applied to the 3D HPE problem as well \cite{YRKS20,SWL21, LWL21,ZZMYCD21}.

In a typical setup, a model processes adjacent video frames to learn temporal representations of human body parts during motion, and then reconstructs a human pose for an interior frame at the inference step. Zheng et al. developed a 3D HPE method called PoseFormer that is based purely on the transformer architecture that encodes and learns both spatial and temporal information \cite{ZZMYCD21}. Li et al. leverages a spatial-temporal framework while finding multiple feasible solutions (given that 3D HPE is an inverse problem) and then leverages a transformer head that aggregates feasible solutions into an optimal solution \cite{li2022mhformer}. The framework operates by generating multiple hypothesis for pose prediction with consequent self-hypothesis refinement and computation of cross-hypothesis interactions. MHFormer achieves superior to previous methods accuracy on MPI-INF-3DHP and Human 3.6M datasets. He et al. developed an Epipolar Transformer to take advantage of 3D data to improve 2D pose estimation, which is challenged in the presence of occlusions and oblique viewing angles \cite{YRKS20}. Shuai introduced a Multi-view and Temporal Fusion transformer to adaptively process varying view numbers and video lengths without calibration \cite{SWL21}.

Even though transformers demonstrate strong capabilities to model complex relationships, they suffer from \textbf{redundancy and over-connectivity}. To address this issue,  researchers in NLP have begun developing different sparsity mechanisms in an attempt to reduce connectivity. For example, Jaszczur et al. proposed a sparse transformer model to scale learning and inference processes \cite{Jaszczur2021SparseIE}. In 3D HPE, Li et al. proposed Strided Transformer \cite{li2022exploiting} to reduce dimensionality of the last linear layers. However, to the best of our knowledge, no work has been done to reduce connectivity in the most "parameter heavy" self-attention mechanism of transformers for the 3D HPE task.

For these reasons, we propose ConvFormer model, which reduces the parameter count extensively relative to \cite{ZZMYCD21}, by an approximate 60 percent. At the spatial level, our multi-scale feature aggregation mechanism is able to capture critical correlations leading to a stronger signal that is more robust relative to \cite{ZZMYCD21}. Moreover, our convolutional self-attention mechanism in the temporal transformer produces queries, keys, and values that extract inter-frame information leading to more diverse attention maps. This enables us to achieve SOTA at a relatively low cost. In the next section, we provide a comprehensive exposition of our methodology.

\section{Method}
In the following subsections, we present an overview of our solution methodology for estimating 3D poses from a sequence of 2D poses, then we describe in our global network architecture, and lastly we present our  dynamic multi-headed convolutional self-attention mechanism.

\subsection{Overview}

\begin{figure*}[!htb]
\centering
\includegraphics[width= \textwidth]{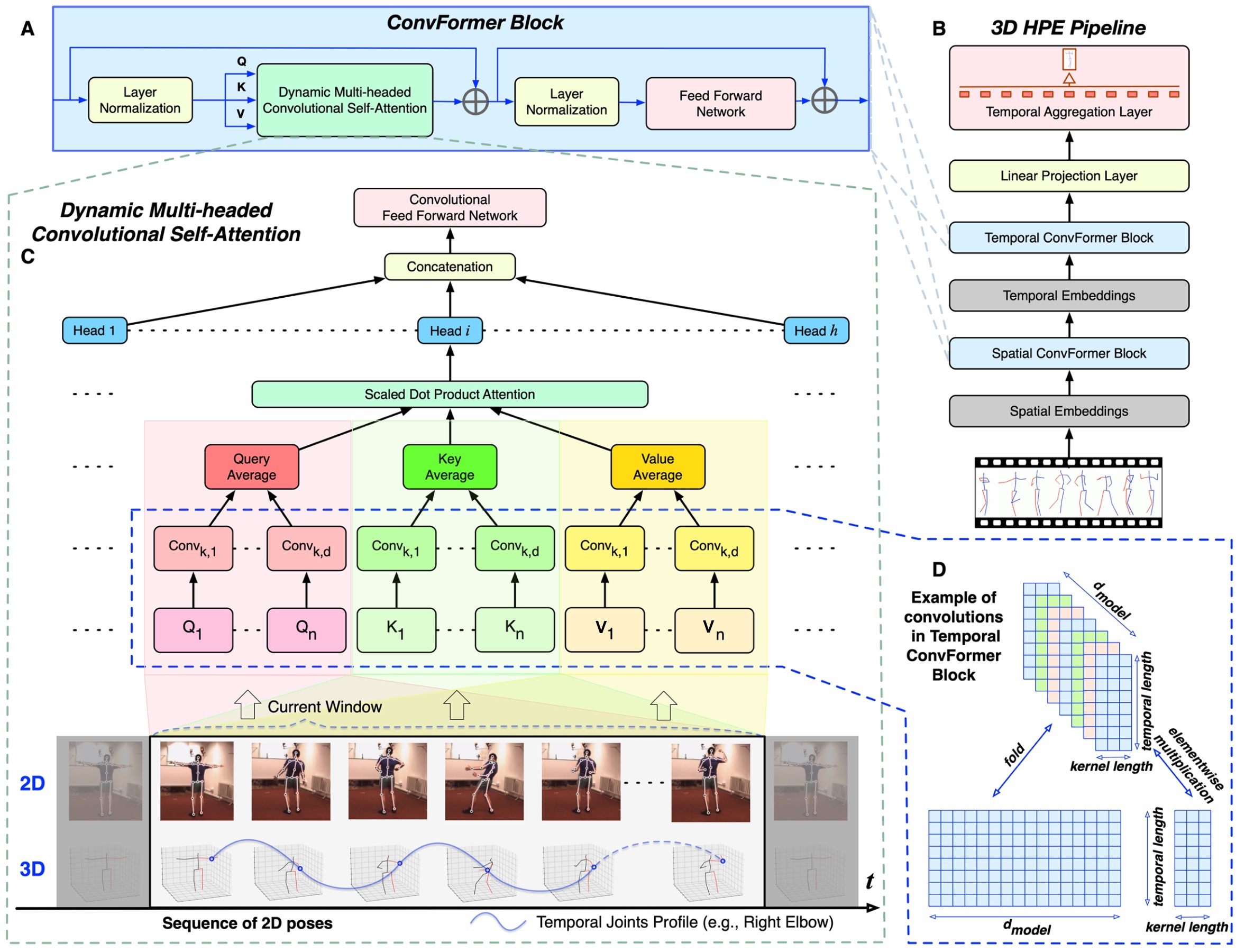}
\caption{\small Panel A depicts  architecture of a ConvFormer Block. Panel B presents the overall pipeline for 3D HPE from a sequence of 2D poses. The central component of a ConvFormer Block is DMHCSA which is depicted in the panel C. A curvy blue line at the bottom of Panel C corresponds to a part of an extracted temporal joints profile of the right elbow joint (for the temporal ConvFormer block). Panel D presents an example of convolutions during generation of Queries, Keys, and Values in a Temporal ConvFormer Block. A filter slides across the feature dimension effectively convolving full temporal profiles of local joint neighborhoods.}
\label{fig: ConvFormer}
\end{figure*}

The overall architecture of our methodology is described in Figure \ref{fig: ConvFormer}. Given a sequence of 2D poses $P = \{P_i\}_{i=1}^{T} \subset \mathbb{R}^{J \times 2}$ where $T$ represents the number of frames in the sequence and J is the number of joints in the skeleton. We seek to reconstruct the 3D poses in the root relative camera reference frame (i.e. the camera reference frame where the root joint sits at the origin). Following \cite{PFGA19}, we predict the 3D pose for the central frame from any such sequence, i.e. $\hat{p}_{\lceil i/2 \rceil} \in \mathbb{R}^{J \times 3}$. Our network contains two Dynamic ConvFormer blocks, one with spatial attention and the other with temporal attention. More specifically, we leverage a spatial attention mechanism to extract frame-wise inter-joint dependencies by analyzing sections of joints that are related. The temporal attention mechanism extracts global inter-frame relationships by analyzing correlations between the temporal profiles of joints. In contrast to \cite{ZZMYCD21}, which queries latent pose representations for individual frames and then computes attention with respect to the temporal axis, our temporal joints profile mechanism fuses temporal information at the querying level prior to computing self-attention with respect to the temporal axis.

\subsection{Network Architecture}
We employ two main components in our network architecture: a \textit{spatial} and a \textit{temporal} ConvFormer. The spatial ConvFormer block extracts a high dimensional feature vector for a single-frames' encoded joint correlations. We assume our input is a 2D pose with $J$ joints that is represented by two coordinates $(u,v)$. Following \cite{ZZMYCD21} we first map the coordinate of each joint into a higher-dimensional feature vector with a trainable linear layer. We then apply a learned positional encoding via summation to retain joint position information. That is, given a sequence of poses $\{P_i\}_{i=1}^{T} \subset \mathbb{R}^{J \times 2}$ and $W \in \mathbb{R}^{2 \times d}$ and $E_{pos} \in  \mathbb{R}^{J \times d}$ we encode $P_i$ as follows:
\begin{equation}
    x_i = P_iW + E_{pos}, \quad i \in \{1,...,T\} \text{.}
\end{equation}
and $d$ represents the dimension of the embedding, $W$ is the trainable linear layer, and $E_{pos}$ is the learned positional encoding. Subsequently, the spatial feature sequence $\{x_i\}_{i=1}^{T}  \subset \mathbb{R}^{J \times d}$ are fed into the spatial ConvFormer which applies the attention mechanism to the joint dimension to integrate information across the complete pose on a per frame basis. $Q,K,V$ are generated via convolutions with weights of the following dimension $(d,d,k)$ where $d$ is the encoded dimension and $k$ is the kernel size and the filter is slid over the joints dimension. The output for the $i$-th frame of the $b$-th spatial ConvFormer block is denoted by $z_i^b \in \mathbb{R}^{J \times d}$ for $i=1,...,T$.

While the spatial ConvFormer seeks to encode correlations between joints in a single frame we leveraged the temporal model to localize sequence wise correlations between the encoded spatial features. This mechanism should be viewed as extracting the temporal profile of a neighborhood of joints, which we call \textbf{\textit{temporal joints profile}} (see Panel D in Figure \ref{fig: ConvFormer}). An early work that leveraged this temporal fusion mechanism was  \cite{KTSLSF14} where Karpathy et al. studied different mechanisms for incorporating temporal information without convolving over the temporal dimension. To further clarify the point, $Q,K,V$ are generated via convolutions with weights of the following dimension $(T, T, k)$ where $k$ is the kernel size and the 1D convolutions have depth the size of input sequence. Thus, one can view our network as fusing into the queries the temporal evolution of a patch of deep joint features immediately. This is very distinct from the temporal attention seen in \cite{ZZMYCD21} which attends complete pose encoding throughout the motion sequence. We note that the output from the spatial ConvFormer block is a sequence $\{z_i^{B}\}_{i=1,...,T} \subset \mathbb{R}^{J \times d}$ where $B$ is the number of spatial blocks and $T$ is the number of frames in the sequence. We note that $z_i^b$ can be represented in $\mathbb{R}^{1 \times J \cdot d}$ and thus concatenate these features along the first axis giving us $X_0 = Concatenate(z_1^B,...,z_T^B) \in \mathbb{R}^{T \times J \cdot d}$. Following this procedure we incorporate a learned temporal embedding to retain information about the deep joint features evolution throughout time, i.e. $E_{temp} \in \mathbb{R}^{T \times J\cdot d}$ and $X = X_0 + E_{temp}$ is the input into our temporal transformer. We note that the output of the $b$-th ConvFormer block with temporal attention is $Z^b \in \mathbb{R}^{T \times J \cdot d}$ where there are $B$ such layers.

Since we follow many-to-one prediction scheme first introduced in \cite{PFGA19} we first down sample the spatial axis with a linear projection and then perform a temporal convolution with one output channel i.e. $\hat{p} = Conv_{T, 1}(Z^BW)$  where $W \in \mathbb{R}^{J \cdot d  \times 3J}$ and $Conv_{T, 1}$ denotes a temporal convolution with one output channel and $T$ input channels.

We trained our network by minimizing the MPJPE (Mean Per Joint Position Error) during optimization. The loss function is defined as
\begin{equation}
    L(p, \hat{p}) = \frac{1}{J} \sum\limits_{i=1}^{J}{\lVert p_i - \hat{p}_i \rVert _2}
\end{equation}
where $p$ is the ground truth 3D pose and $\hat{p}$ is the predicted pose and $i$ is indexing specific joints in the skeleton.

\subsection{Dynamic Multi-Headed Convolutional Self-Attention}
A core novelty of this paper is the dynamic multi-headed convolutional self-attention mechanism. This is introduced to reduce the over connectedness witnessed in classic transformer architectures while simultaneously extracting contexts at different scales. An additional novelty is the type of representations being queried in our temporal ConvFormer block. Instead of generating queries, keys, and values, that are latent pose representations for individual frames and attending the temporal axis; we query temporal joints profiles effectively fusing temporal information prior to the attention mechanism.

Convolutional Scaled Dot Product Attention can be described as a mapping function that maps a query matrix Q, a key matrix K, and a value matrix V to an output attention matrix -- where the matrix entries are scores representing the strength of correlation between any two elements in the dimension being attended. We note that $Q,K,V \in \mathbb{R}^{N \times d}$ where N is the length of the sequence and d is the dimension. In our Spatial ConvFormer $N = J$ and in the Temporal ConvFormer $N = T$. The output of the scaled dot product attention can be expressed as

\begin{equation}
    Attention(Q,K,V) = Softmax(QK^T / \sqrt{d})V  \text{.}
\end{equation}

The query, keys, and values, are computed in the same manner for a fixed filter length.  We demonstrate how $Q$ can be generated, and note that $K$ and $V$ are computed in an identical manner.

\begin{equation}
    Q = Conv_{n, d_{out}}(z) = \sum\limits_{i=1}^{d_{in}}\sum\limits_{k=1}^{\kappa}{w_{d_{out}, i, k} \cdot  z_{i, n - \frac{\kappa-1}{2} +k}}
    \label{eqn:somelabel}
\end{equation}
Here, $\kappa$ denotes the kernel size and $d_{out}$ denotes output dimension. This is juxtaposed  against the classic scaled dot product attention introduced in \cite{VSPUJGKP17} where queries, keys, and values are generated via a linear projection 
\begin{equation}
    Q = W_Q z \quad K = W_K z \quad V = W_V z
\end{equation}
which provides global scope but causes redundancy due to the complete connectivity. In our dynamic convolutional attention mechanism we introduce sparsity via convolutions to decrease connectivity while simultaneously fusing complete temporal information prior to the scaled-dot-product-attention. ConvFormers' ability to provide context at different scales is attributable to the dynamic feature aggregation method. Moreover, due to our convolution mechanism we query on inter-frame level where we learn the temporal joints profile. 
To this end, we use $n$ convolutional filter sizes to extract different local contexts at scales $\{\kappa_i\}_{i=1}^{n}$ and then perform an averaging operation to generate the final query, keys, and values that we apply attention to, following ideas presented in \cite{ZKLOT16}:
\begin{equation}
\begin{aligned}
    Q = Concat(Q_1,...,Q_n) \eta_Q =  \sum\limits_{i=1}^{n}{\eta_Q(i)Q_{i}} \\
     \text{where} \quad \sum\limits_{i=1}^{n}{\eta_Q(i)} = 1 \\
\end{aligned}
\end{equation}
where $n$ is the number of convolution filters used, $\eta_Q \in \mathbb{R}^{n \times 1}$ is a learned parameter and $Q_i$ are generated as in equation \ref{eqn:somelabel}.

Dynamic Multi-headed Convolutional Self-Attention (DMHCSA) leverages  multiple heads to jointly model information from multiple representation spaces. As seen in Figure \ref{fig: ConvFormer} each head applies scaled dot-product self-attention in parallel. The output of the DMHCSA block is the concatenation of $h$ attention head outputs fed into a feed-forward network.
\begin{equation}
\begin{aligned}
    DMHCSA(Q,K,V) = Concatenate(H_1,...,H_h) \\
    \text{where} \quad  H_i = Attention(Q_i,K_i,V_i), \quad i \in \{1,...,h\}
\end{aligned}
\end{equation}
where  $Q_i$, $K_i$, and $V_i$ are computed via the procedure defined above.

Then the ConvFormer block is defined by the following equations:
\begin{equation}
\begin{aligned}
    X^{'}_{b} = DMHCSA(LN(X_{b-1})) +  X_{b-1}, \quad b = 1,...,B \\
    X_{b} =  FFN(LN(X_{b}^{'})) + X_b^{'}, \quad b = 1,...,B
\end{aligned}
\end{equation}
where $LN(\cdot)$ denotes layer normalization same as \cite{KDWHUBMDHGUZ21,YCWYSJTFY21}. and $FFN$ denotes a feed forward network. Both the spatial and temporal ConvFormer blocks consist of $B_{sp}$ and $B_{temp}$ identical blocks. The output of the spatial ConvFormer encoder is $Y \in \mathbb{R}^{T \times J \times d}$ where  $T$ is the frame sequence length, J is the number of joints, and $d$ is the embedding dimension. The output of the temporal ConvFormer is $Y \in \mathbb{R}^{T \times Jd}$.

\section{Experiments}

\subsection{Datasets and Evaluation Protocols}
Our proposed method is evaluated on three common datasets: Human3.6M \cite{IPOS13}, HumanEva \cite{SBB10}, and MPI-INF-3DHP \cite{mono-3dhp2017}. Human3.6M consists of approximately 2.3 million images from 4 synchronized video cameras capturing video at 50 Hz. There are 7 subjects performing 15 distinct actions and each action is performed twice per subject. We train on subjects (S1, S5, S6, S7, S8) and validate on subjects (S9, S11) following previous works \cite{CFSZCL21,LL19,mono-3dhp2017,ZZMYCD21,li2022mhformer}. 
We evaluate our method on H36M under three different protocols. The mean per joint position error (MPJPE) which is referred to as Protocol I in many works \cite{FXWLZ18,KKA19,PFGA19}. Procrustes analysis or rigid alignment denoted by P-MPJPE is calculated as the Euclidean distance between the ground-truth and the optimal $SE(3)$  transformation aligning the predicted pose with the ground-truth. This is referred to  as Protocol II as in \cite{MHRL17,RL18}. Lastly, we  evaluate temporal smoothness via the mean per joint velocity error, referred to as MPJVE (the mean across joints of the finite difference velocity approximations) or Protocol III as in \cite{PFGA19,WYXL20}. HumanEva on the other hand is a much smaller dataset with less than 50k frames and only 3 subjects (S1, S2, S3) performing three actions. We evaluate our method with respect to Protocol II following previous works  (e.g. \cite{PFGA19}. Lastly, we evaluated on MPI-INF-3DHP to assess our model's generalizability. MPI consists of roughly 1.3 million frames. This dataset contains more diverse motions than the previous two datasets. Following the setting in \cite{CFSZCL21,LL19,mono-3dhp2017,ZZMYCD21,li2022mhformer} we report the following metrics: MPJPE, Percentage of Correct Keypoint (PCK) with the threshold of 150mm, and Area Under Curve (AUC) for a range of PCK thresholds.

\subsection{Implementation Details}
We implemented our proposed solution methodology with PyTorch \cite{PGCCYDLDAL17} and trained using two NVIDIA RTX 3090 GPUs. We trained on H3.6M using 5 different frame sequence lengths when conducting our experiments, $T=9, 27, 81, 143, 243$. Following \cite{PFGA19} we augment our datasets with flipping poses horizontally. We train our models for 60 epochs with an initial learning rate of  $1e-3$ and a weight decay factor of $0.95$ after each epoch. We set the batch size to $1024$ and utilize  stochastic depth \cite{HSLSW16} of $0.2$. We also use a dropout \cite{JMLR:v15:srivastava14a} rate of $0.2$ on the dynamic feature aggregation inside of the convolutional self-attention mechanism. We benchmark on H3.6M using both CPN \cite{CWPZY18} detections following \cite{PFGA19,CFSZCL21,WYXL20,ZZMYCD21} and ground-truth 2D poses. Furthermore, we benchmark on HumanEva using three different frame sequence lengths of $T=9$, $T=27$, and $T=43$ following \cite{LSWCCA20}. Lastly, following \cite{ZZMYCD21,li2022mhformer} we further assess the generalization ability of our solution methodology on MPI-INF-3DHP dataset. We use 2D pose sequences of length $T=9$ as our model input and we evaluate using three metrics, percentage of correct keypoints (PCK), area under the curve (AUC), and MPJPE.

\section{Results and Discussion}

\begin{figure*}[]
    \centering
    \subfigure[a]{\includegraphics[width=0.45\textwidth]{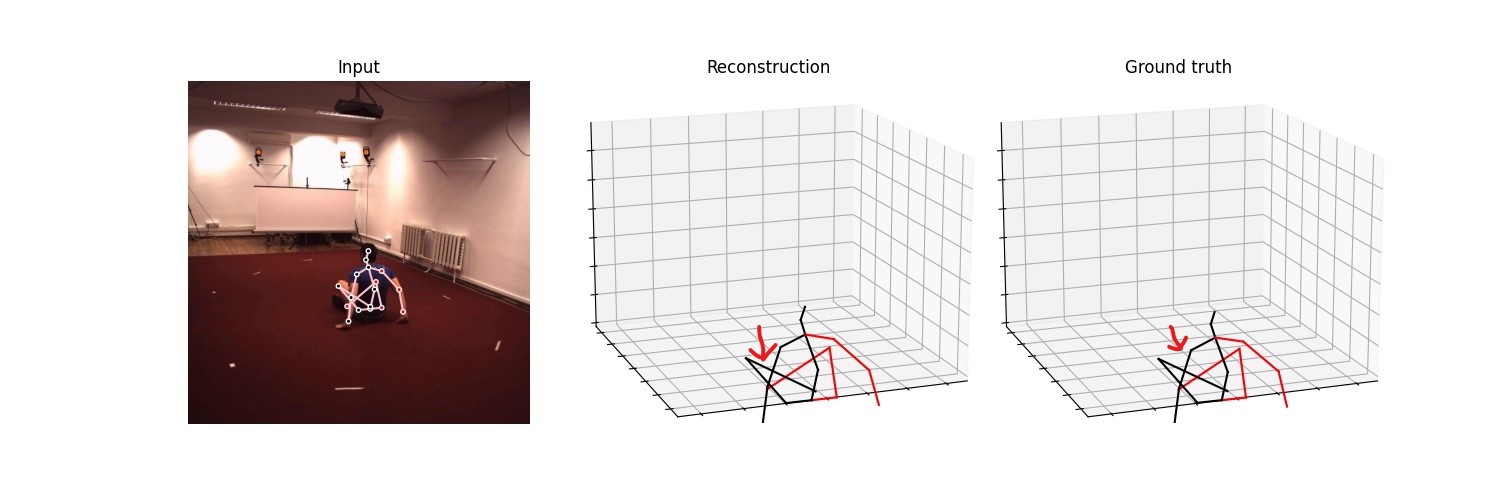}}
    \subfigure[b]{\includegraphics[width=0.45\textwidth]{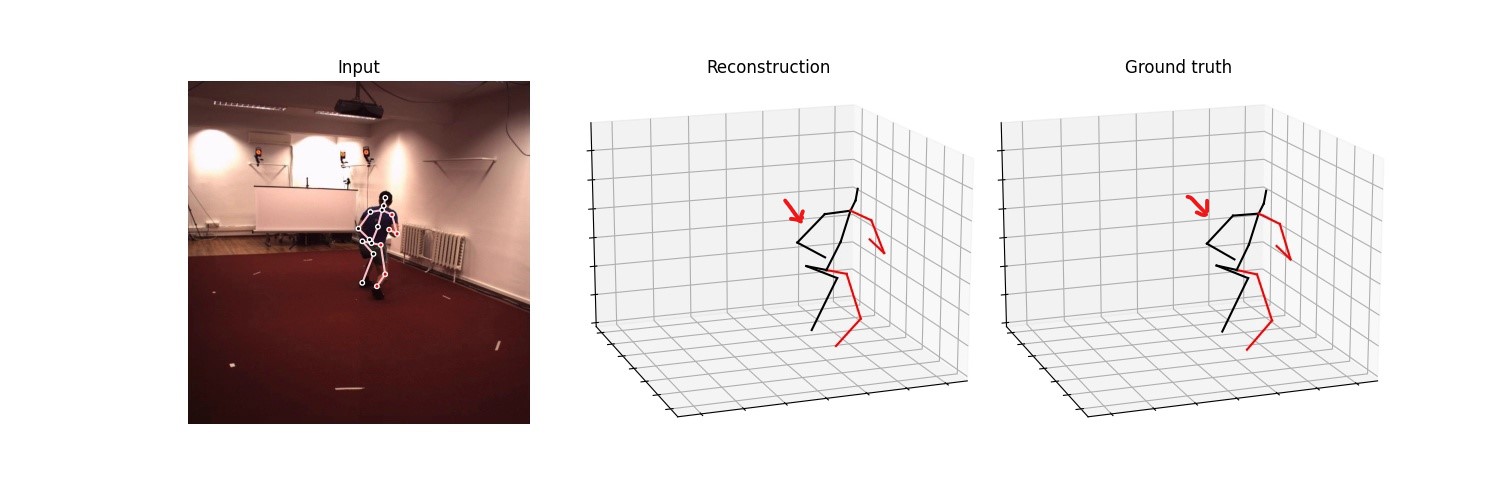}}
    \subfigure[c]{\includegraphics[width=0.45\textwidth]{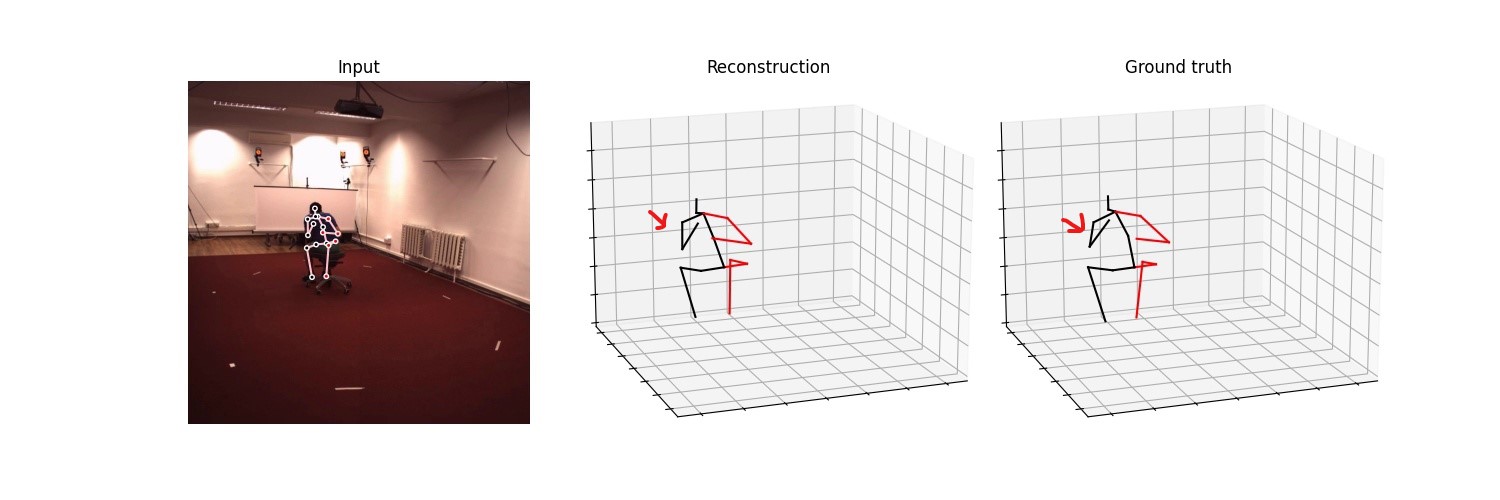}}
    \subfigure[d]{\includegraphics[width=0.45\textwidth]{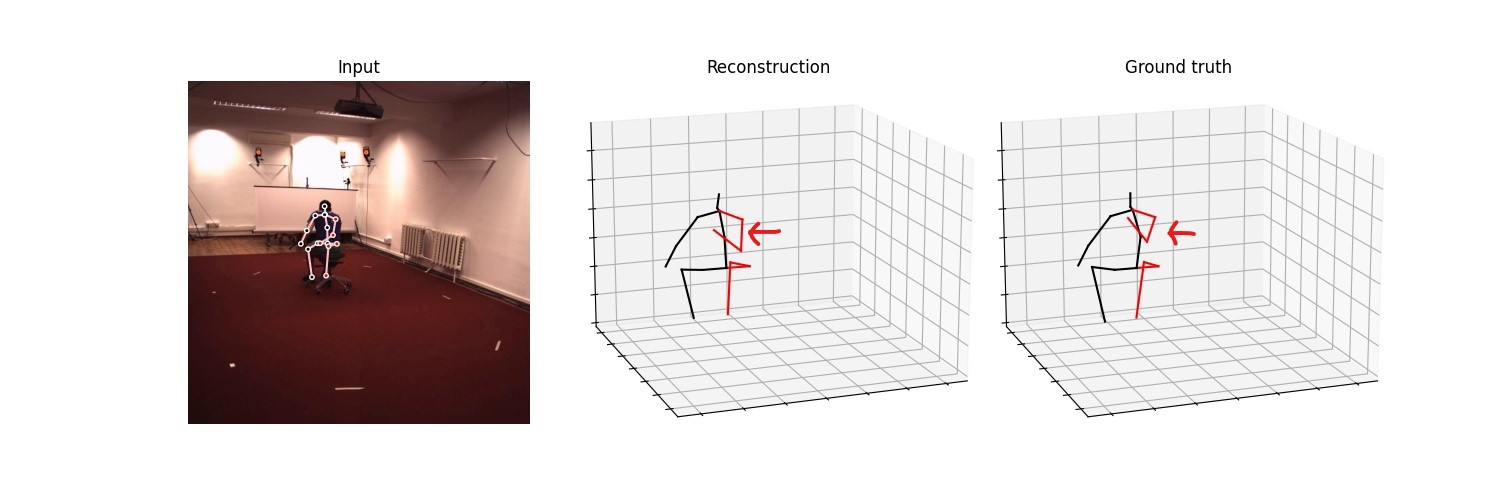}}
    \caption{\small Qualitative examples of S11 from H36M displaying ConvFormer's effectiveness: (a) Sitting Down action with heavy occlusion on lower extremities, (b) demonstrates high quality reconstruction in the presence of slight occlusions, (c) heavy occlusion from camera and ConvFormer still captures the correct pose from previous frame information (d) slight failure case in presence of  occlusion from right arm.}
    \label{fig:H36MExamples}
\end{figure*}

\begin{figure*}[]
\centering
    \subfigure{\includegraphics[width=0.22\textwidth]{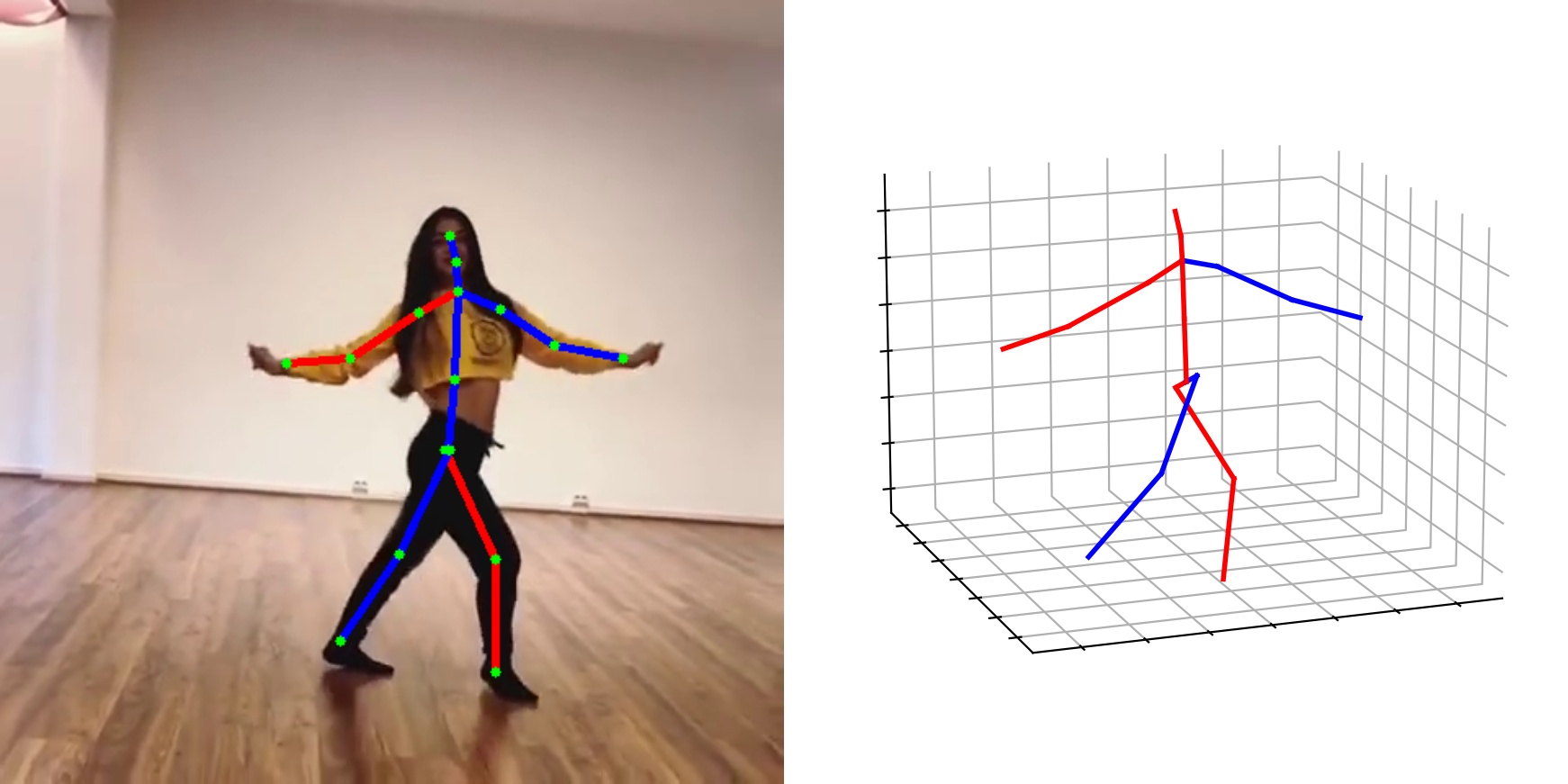}} 
    \subfigure{\includegraphics[width=0.22\textwidth]{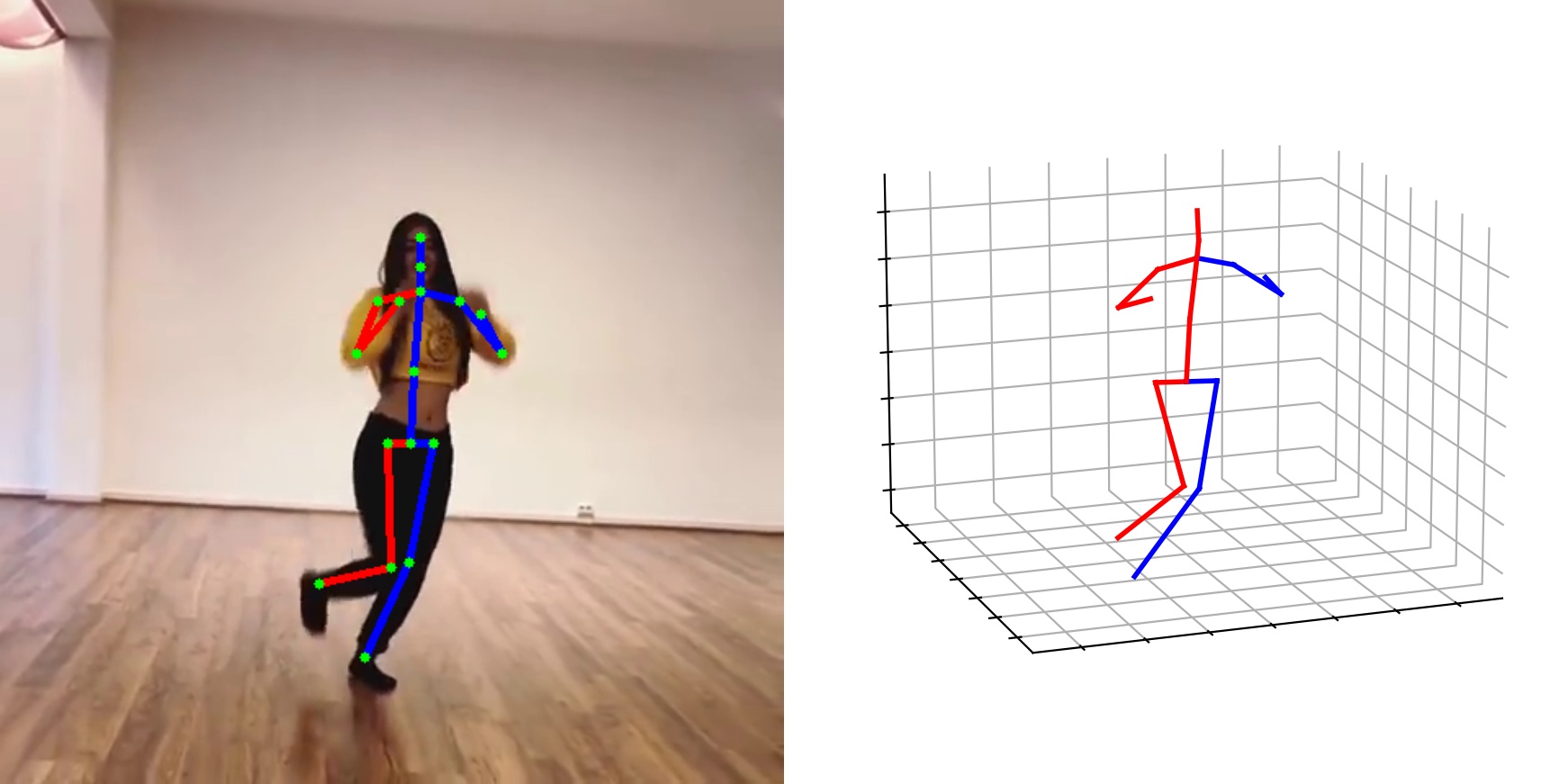}} 
    \subfigure{\includegraphics[width=0.22\textwidth]{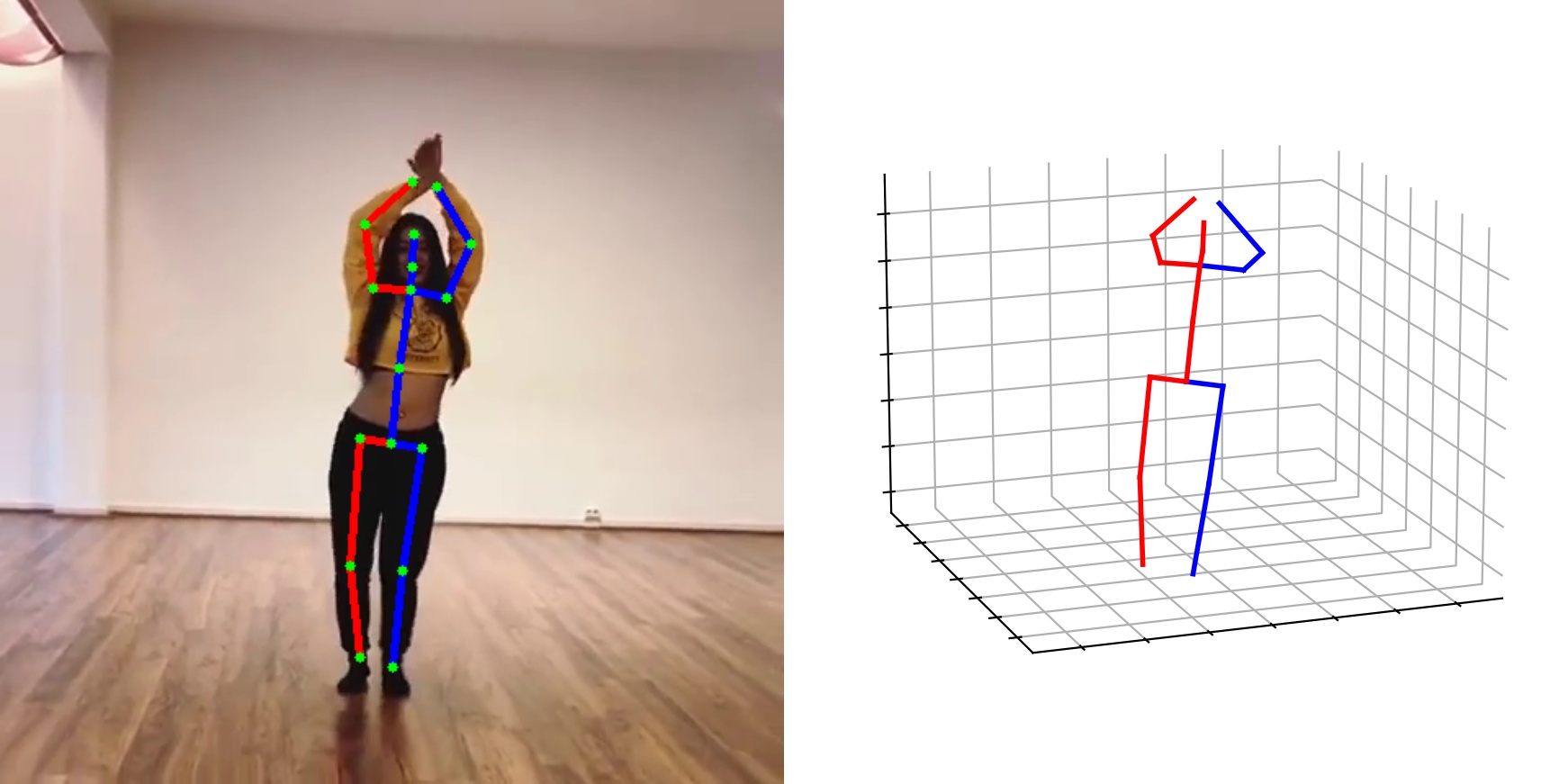}}
        \subfigure{\includegraphics[width=0.22\textwidth]{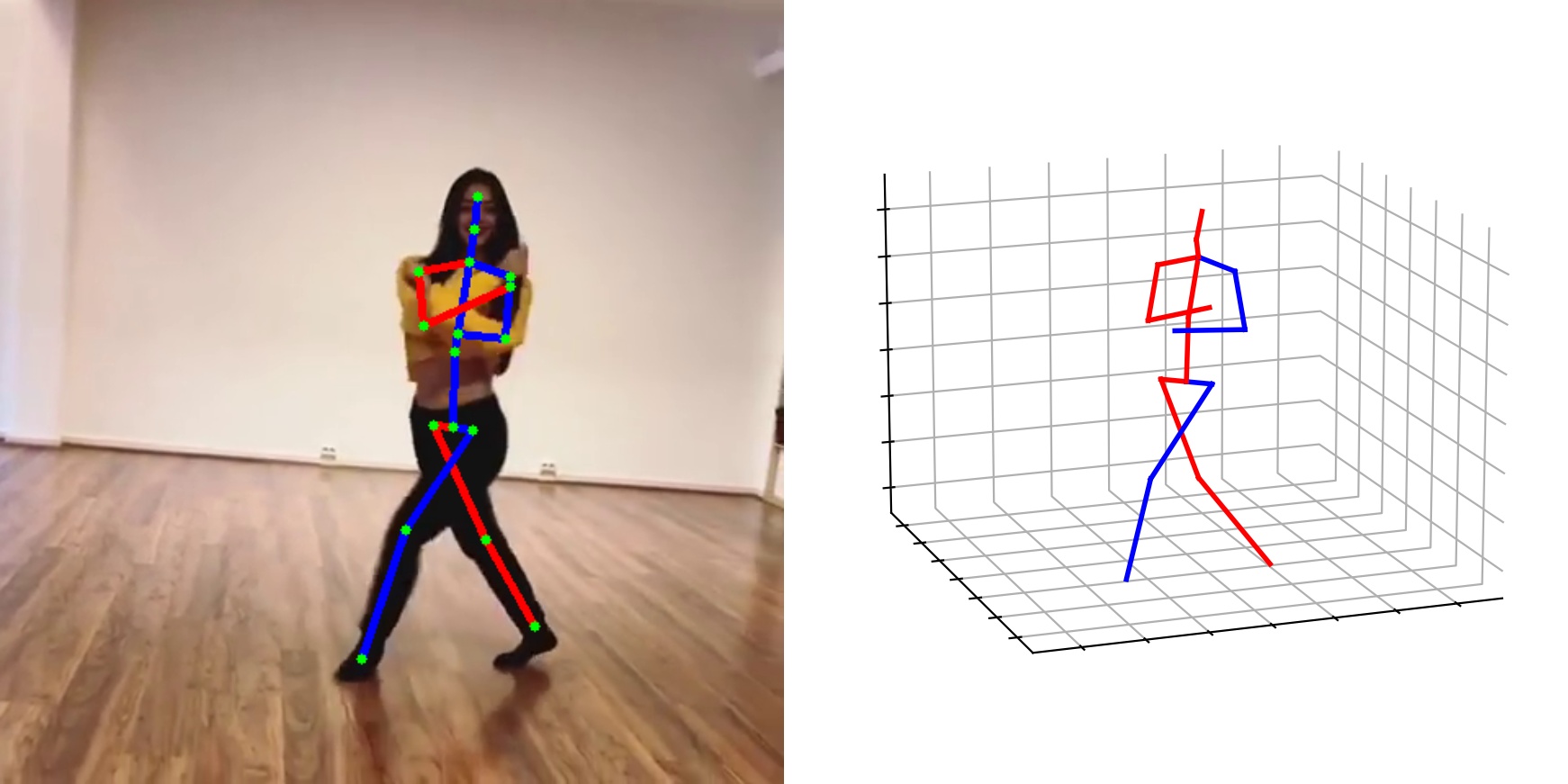}}
        
        \subfigure{\includegraphics[width=0.22\textwidth]{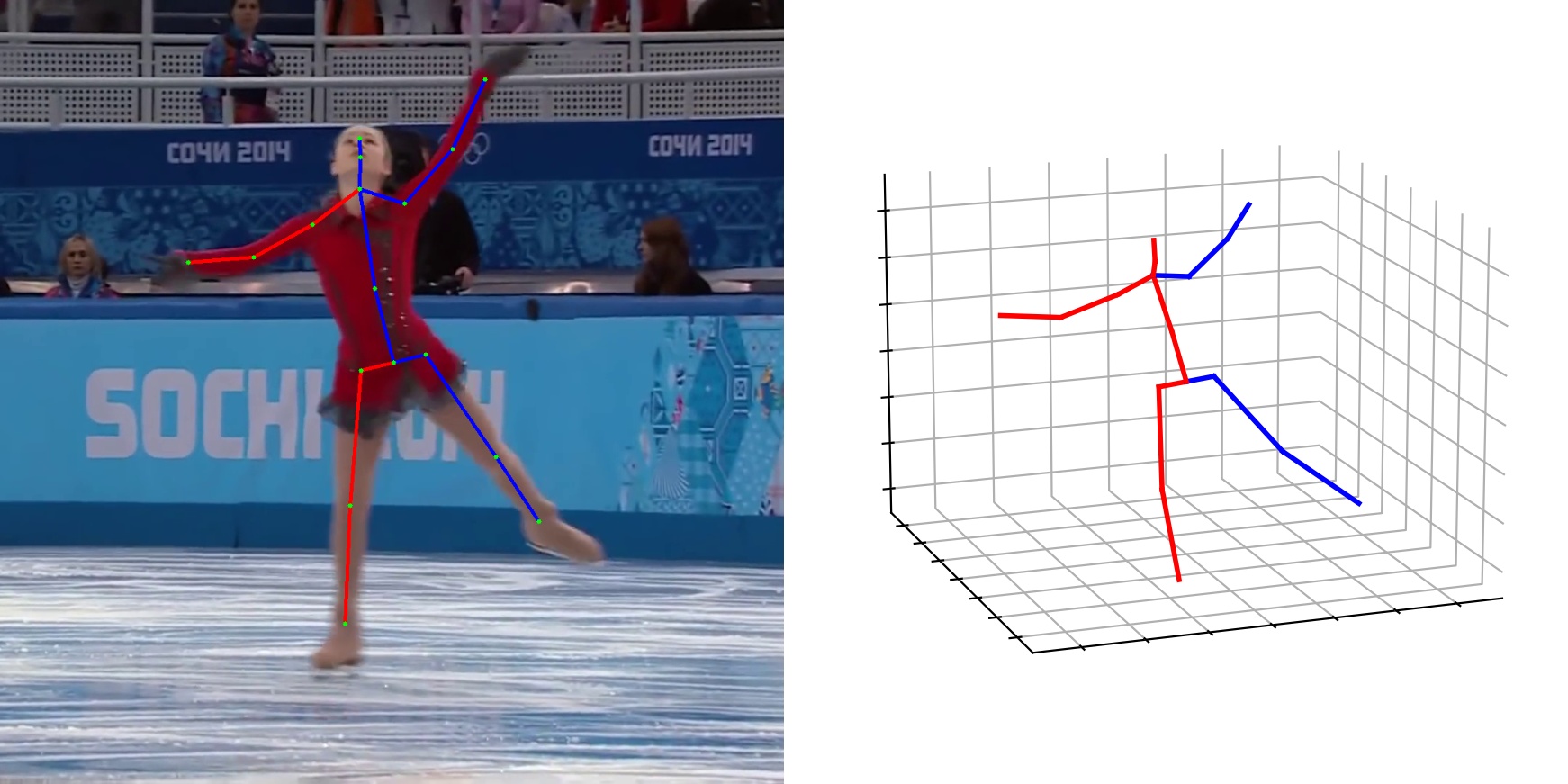}} 
    \subfigure{\includegraphics[width=0.22\textwidth]{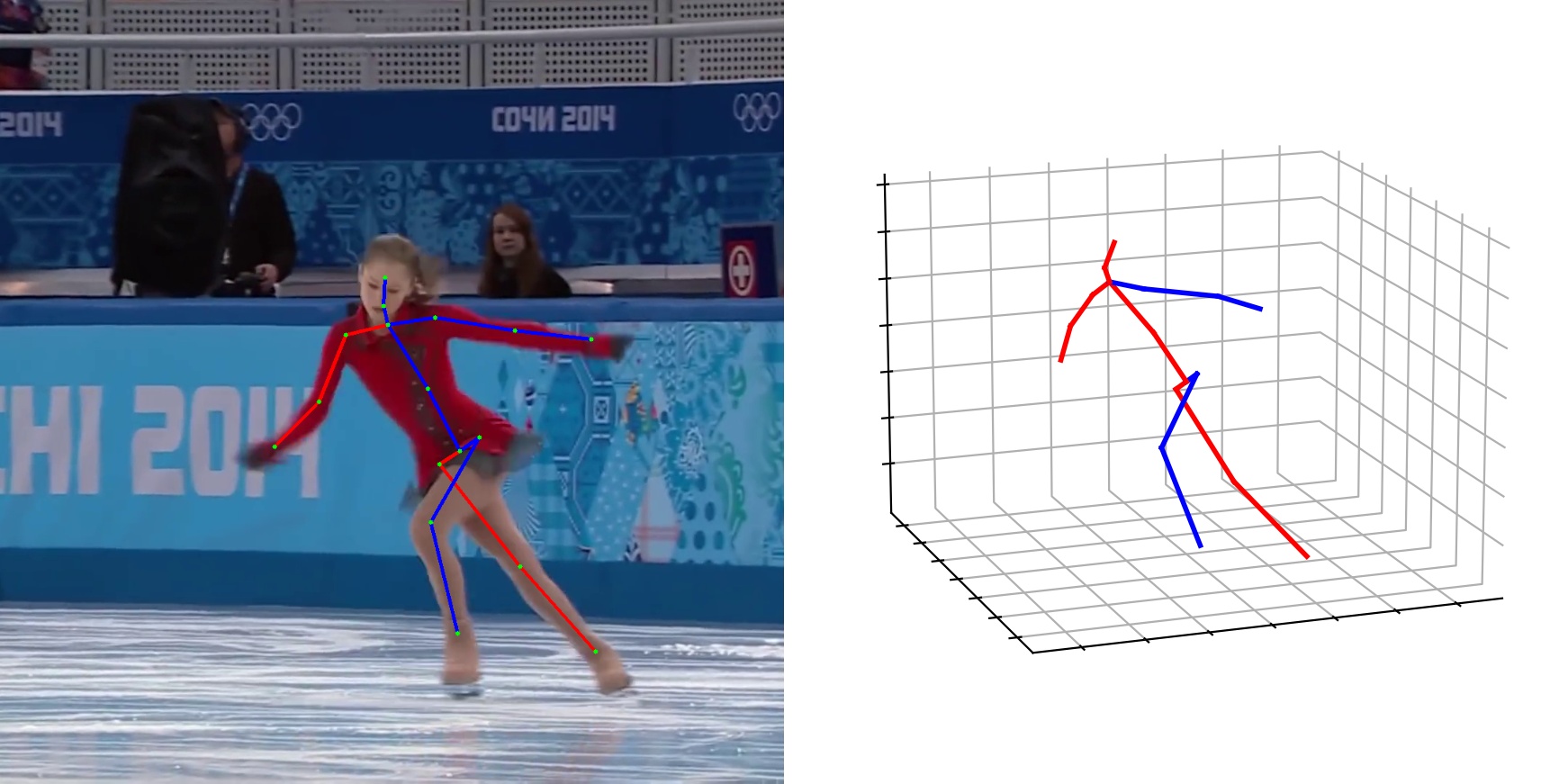}} 
    \subfigure{\includegraphics[width=0.22\textwidth]{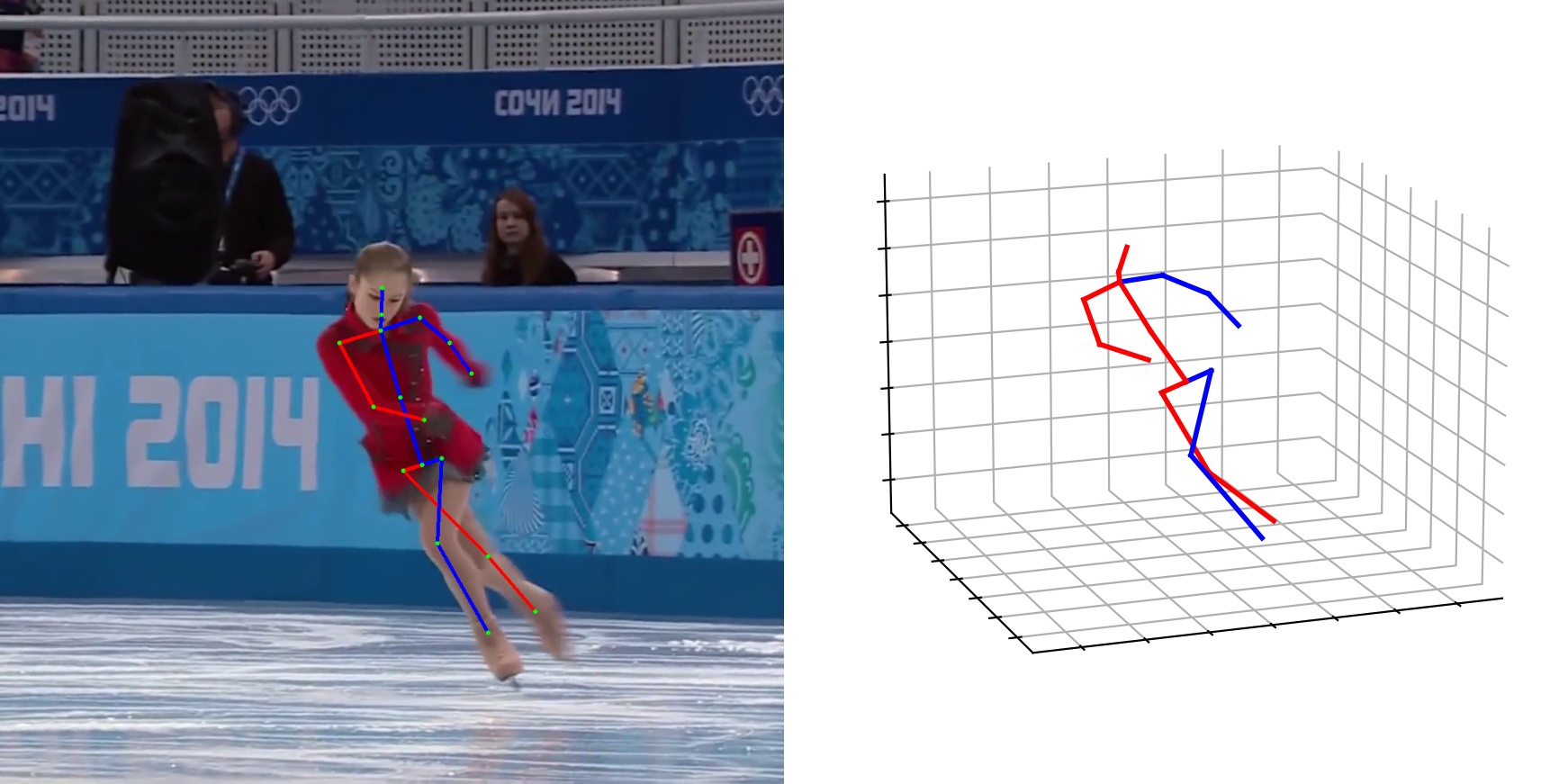}}
        \subfigure{\includegraphics[width=0.22\textwidth]{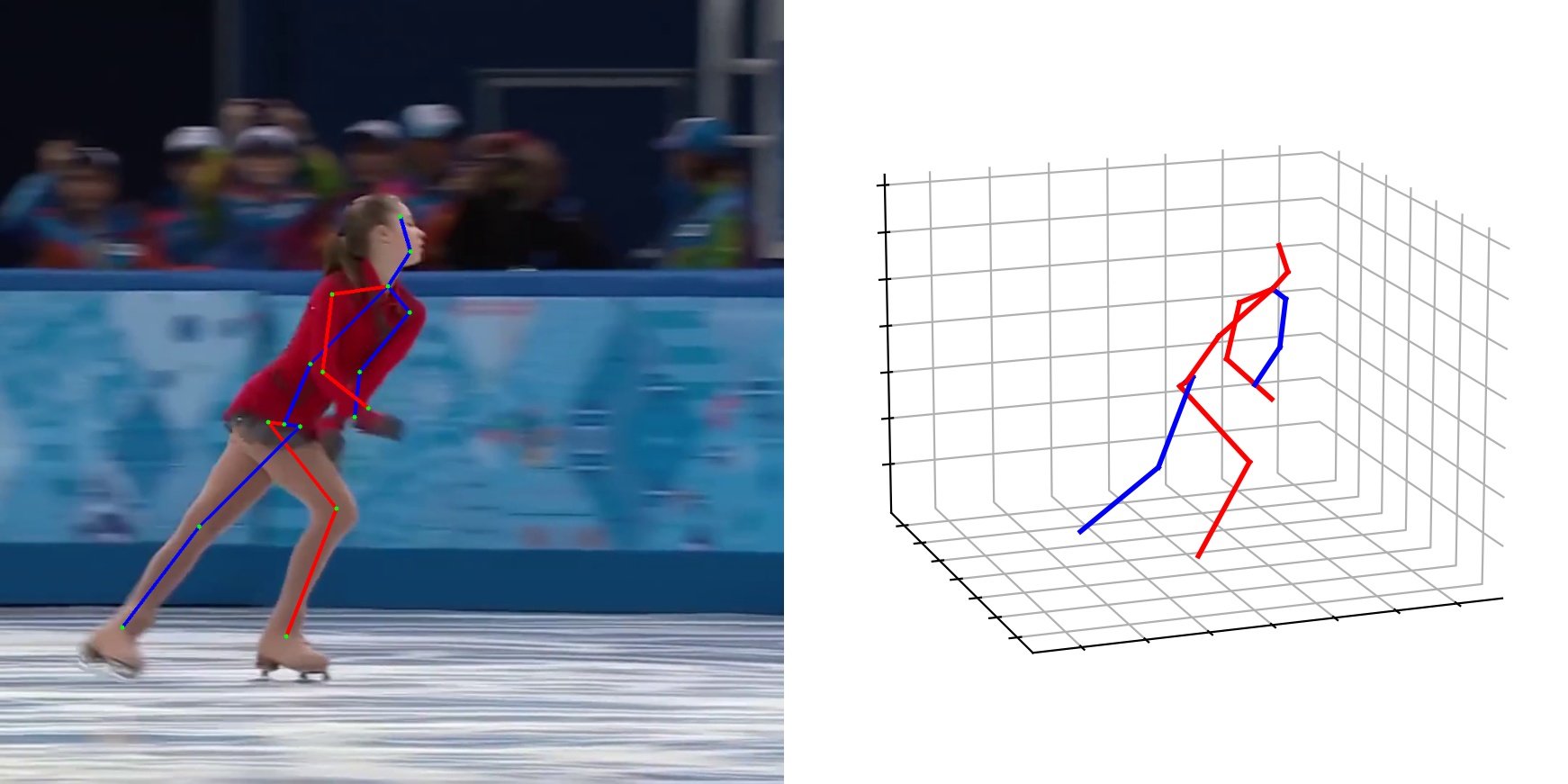}}
        
    \subfigure{\includegraphics[width=0.22\textwidth]{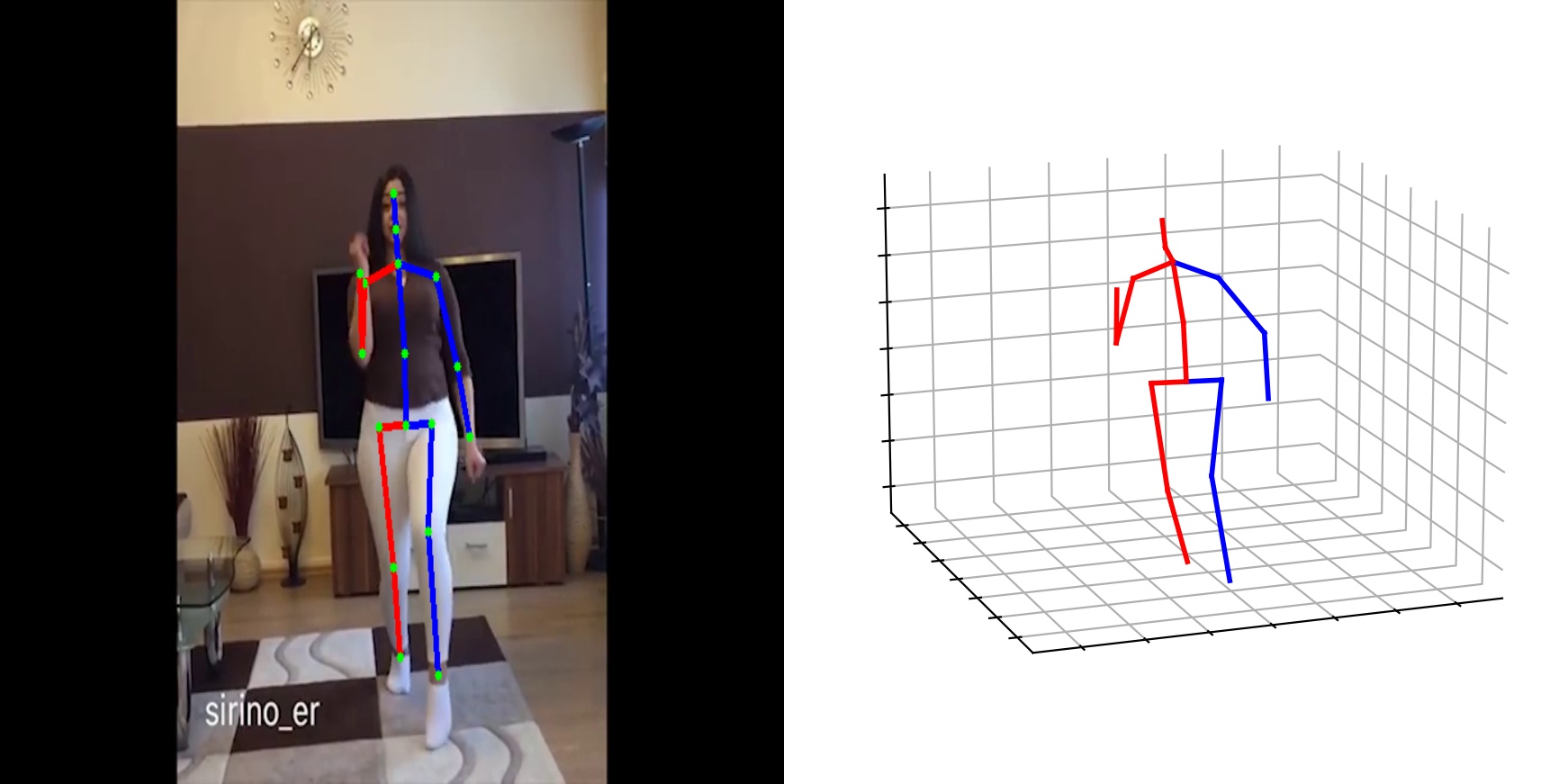}}
        \subfigure{\includegraphics[width=0.22\textwidth]{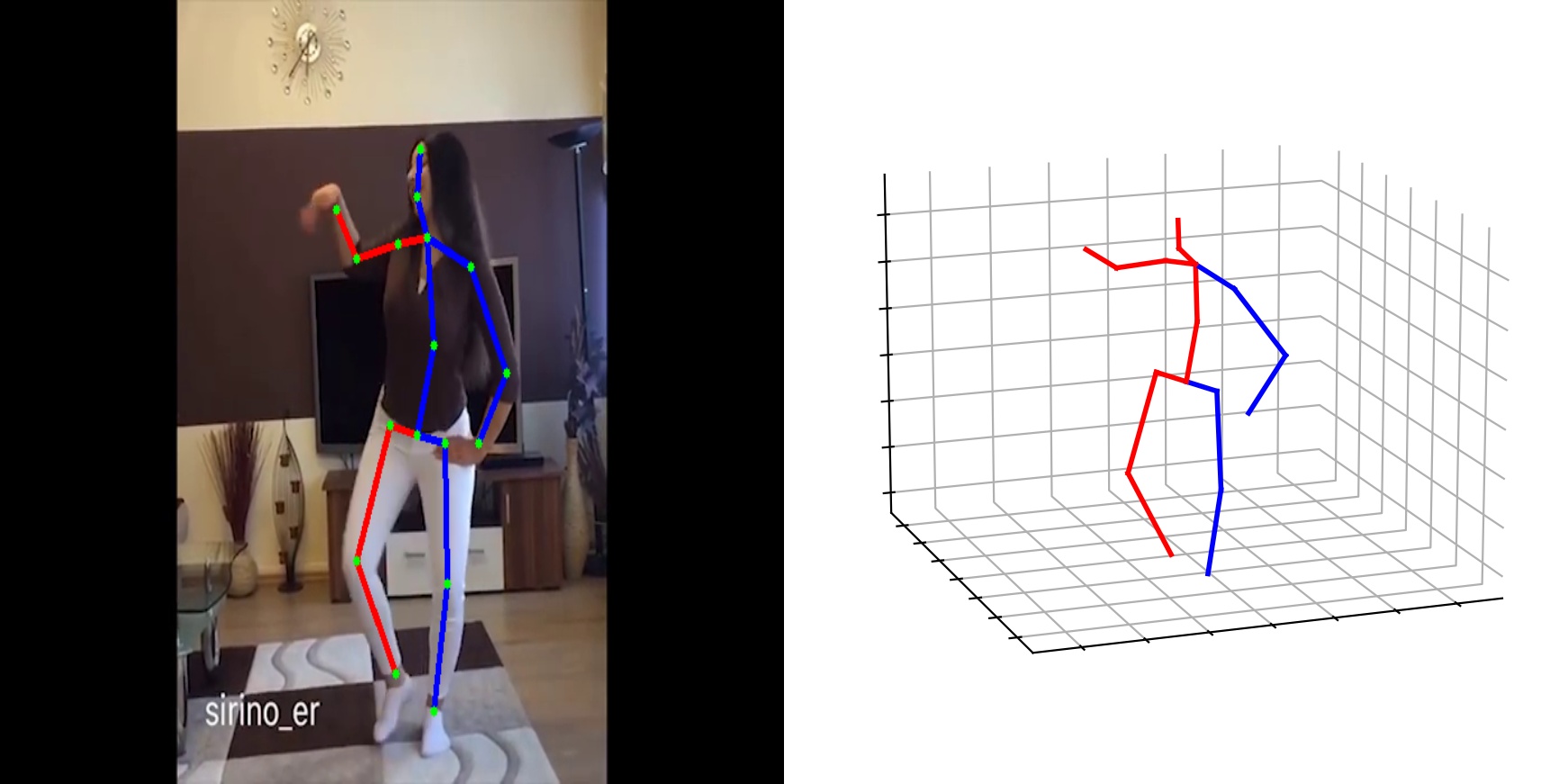}}
    \subfigure{\includegraphics[width=0.22\textwidth]{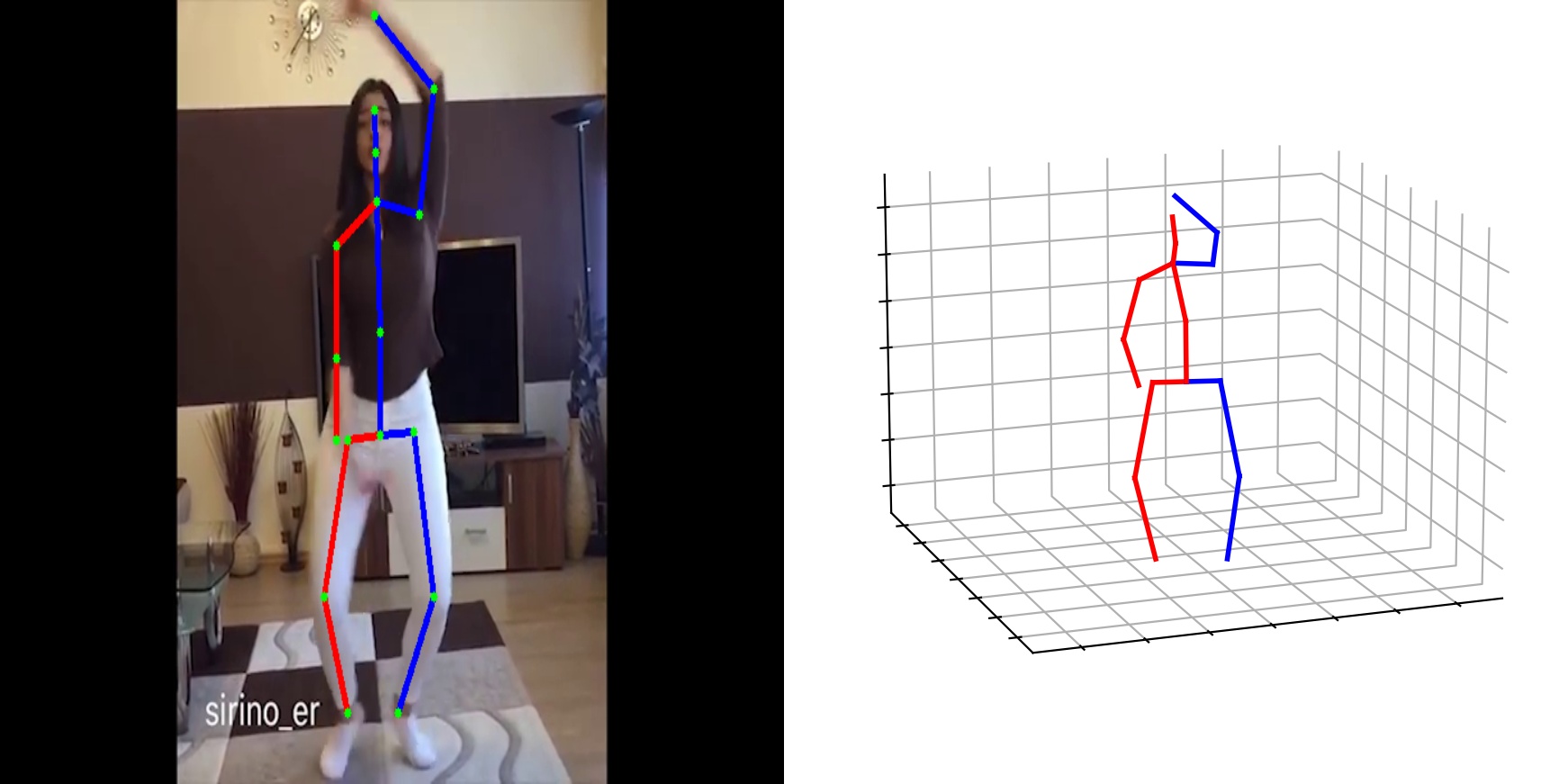}}
        \subfigure{\includegraphics[width=0.22\textwidth]{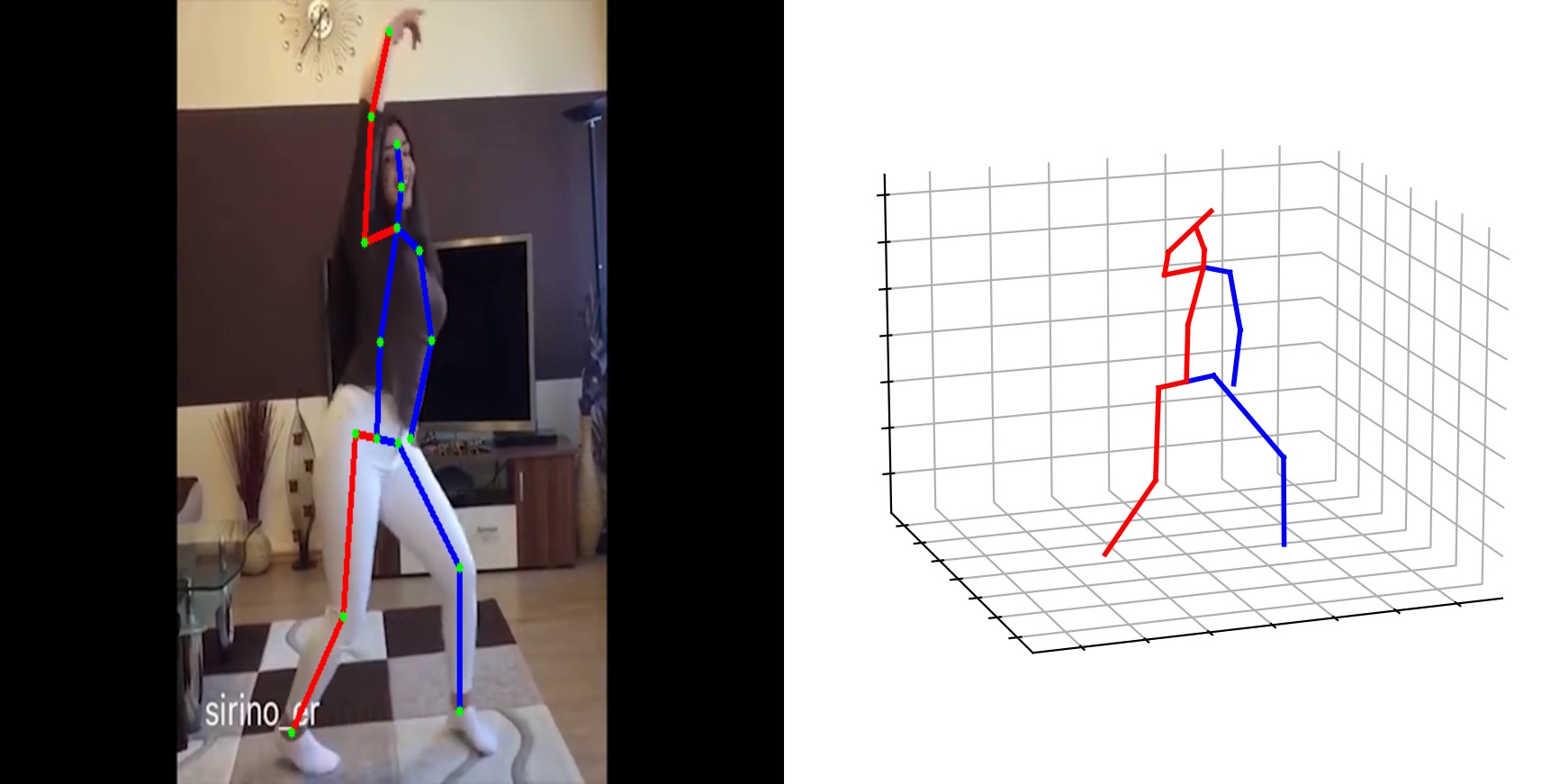}}
        
        \subfigure{\includegraphics[width=0.22\textwidth]{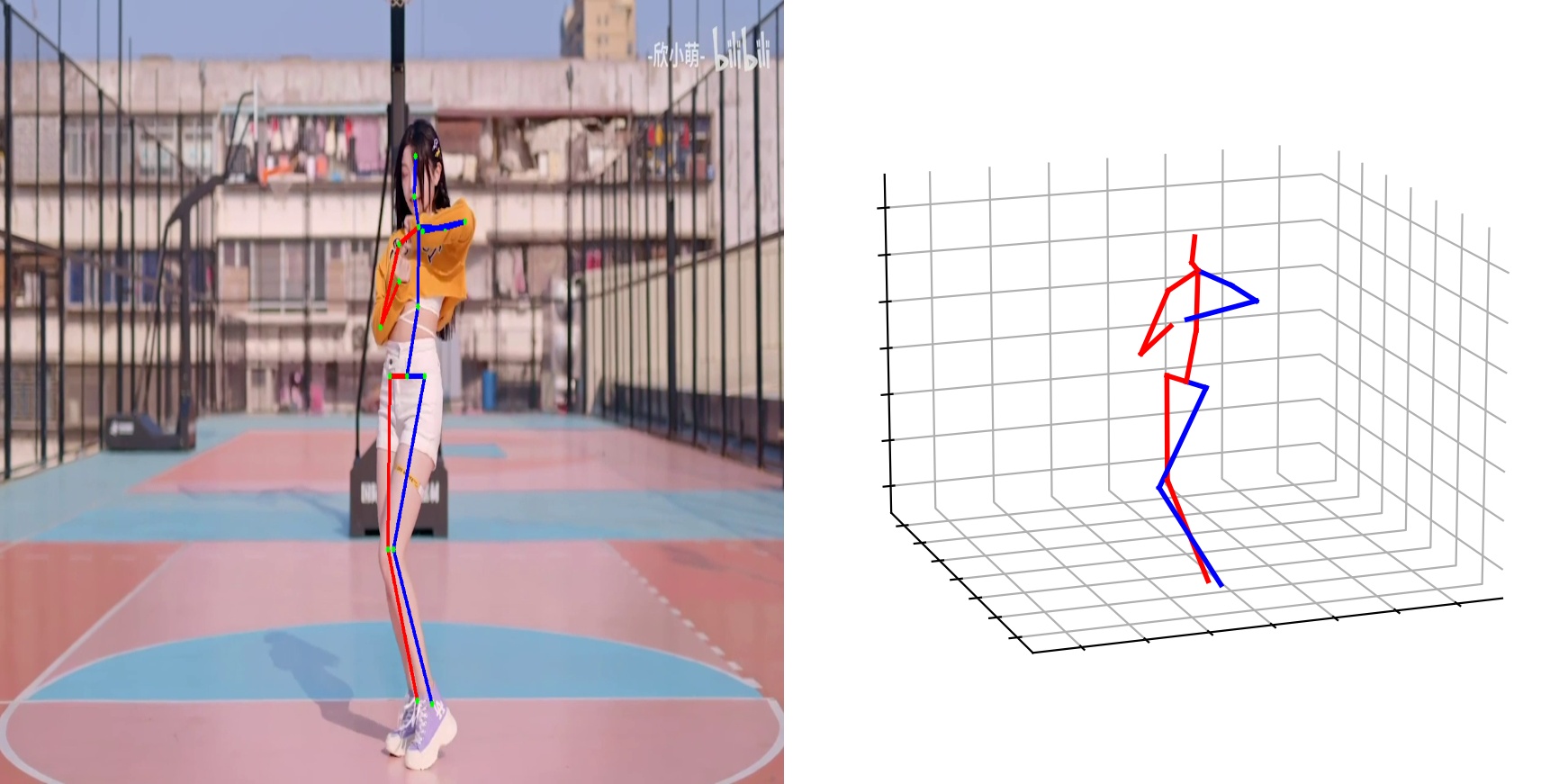}} 
    \subfigure{\includegraphics[width=0.22\textwidth]{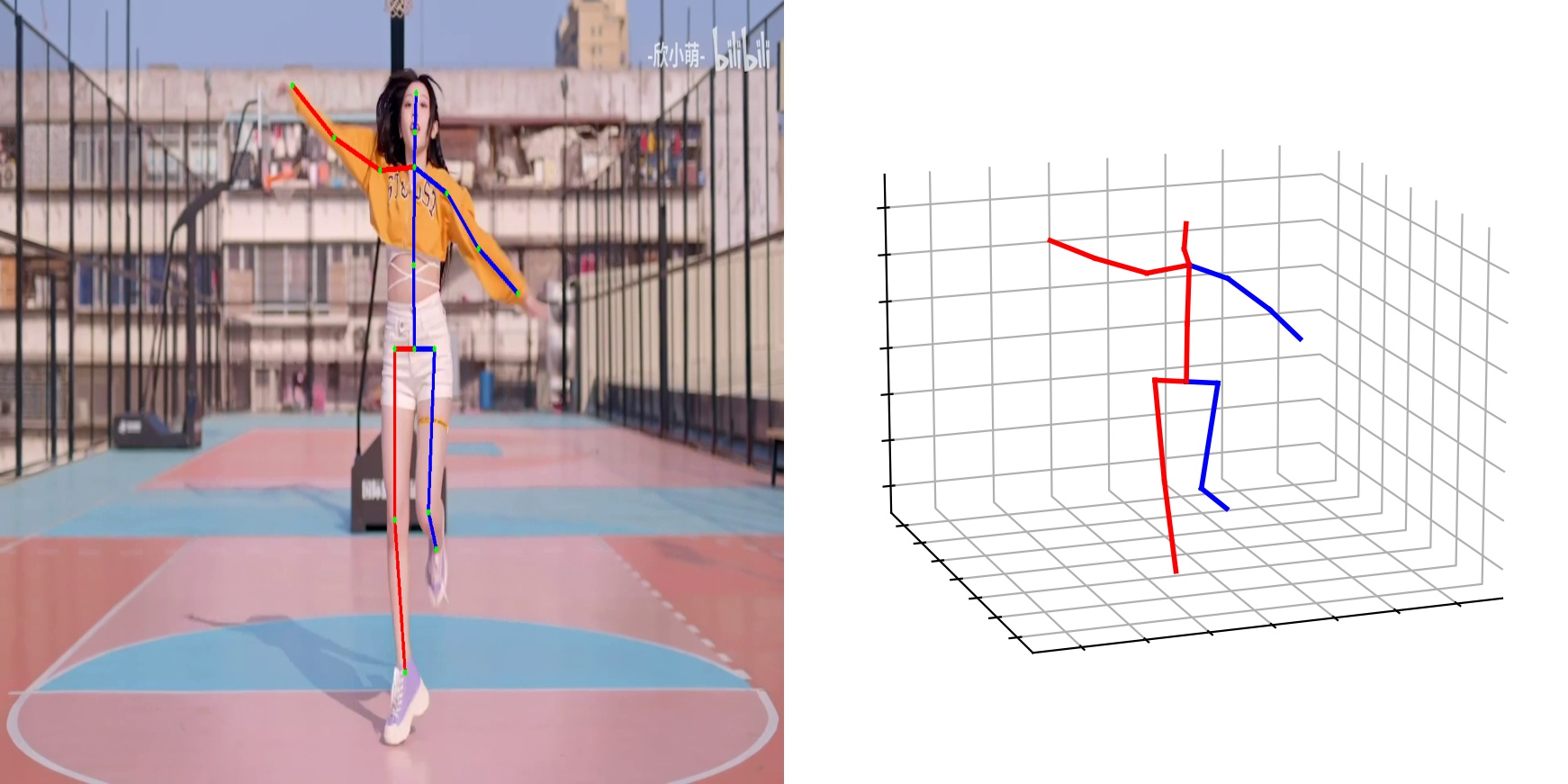}} 
    \subfigure{\includegraphics[width=0.22\textwidth]{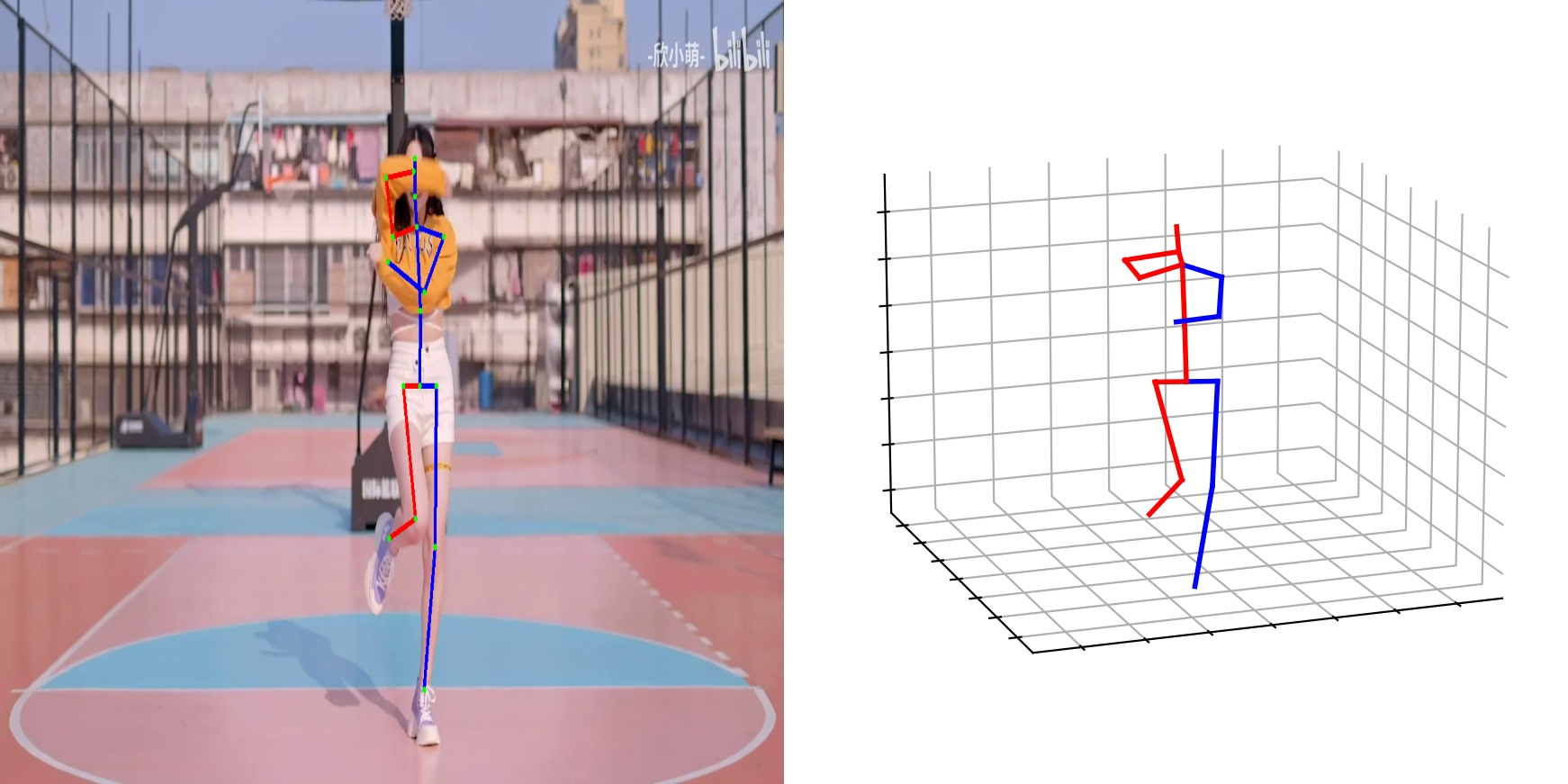}}
        \subfigure{\includegraphics[width=0.22\textwidth]{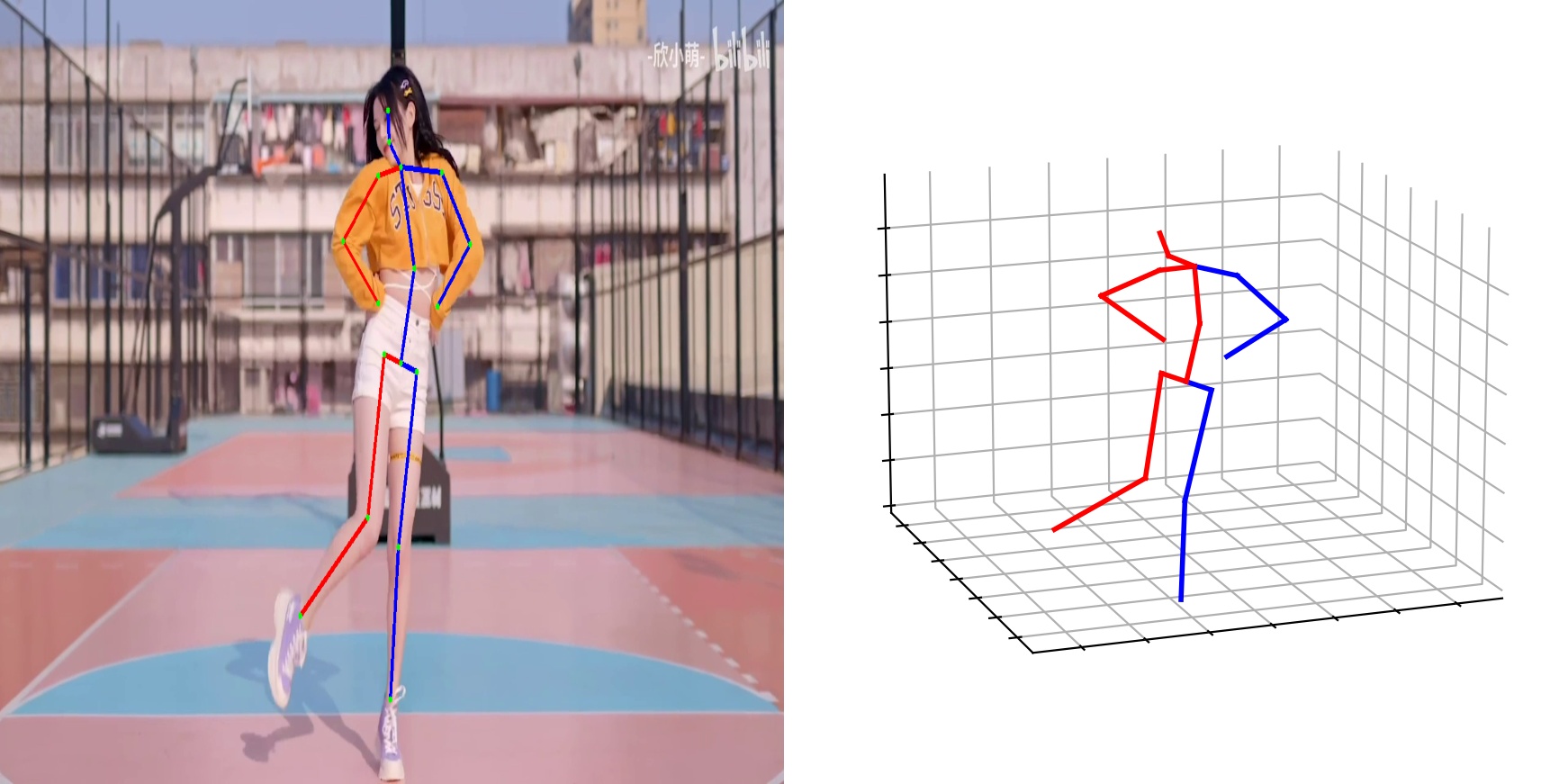}}
    \caption{\small Qualitative results for ConvFormer on challenging In-The-Wild videos. }
    \label{fig:wild}
\end{figure*}

\subsection{Comparison with State-of-the-Art}

\begin{table*}[!htb]

\centering
\caption{\small The first block reports MPJPE for GT-inputs and the second block is MPJPE for CPN detections. The third block reports P-MPJPE for CPN detections. The fourth block reports MPJPV for CPN detections and the fifth is MPJPV for GT-inputs. 
Best is in \textcolor{red}{Red} and second is in \textcolor{blue}{Blue}.}

 \resizebox{\columnwidth}{!}{\begin{tabular}{*{17}{|c}|}  
 
  \hline
  \textbf{GT-MPJPE (mm)} &\textbf{Dir.} & \textbf{Disc.} & \textbf{Eat} & \textbf{Greet} & \textbf{Phone} & \textbf{Photo} & \textbf{Pose}  & \textbf{Purch.} & \textbf{Sit} & \textbf{SitD.} & \textbf{Smoke} & \textbf{Wait} & \textbf{WalkD.} & \textbf{Walk}  & \textbf{WalkT.} & \textbf{Avg.} \\
 \hline
  Hossain and Little \cite{RL18}  & 35.2 & 40.8 & 37.2 & 37.4 & 43.2 & 44.0 &  38.9 &  35.6 &  42.3 & 44.6 & 39.7 & 39.7 & 40.2 & 32.8 &35.5 & 39.2  \\
  Pavllo et al. \cite{PFGA19} & -- & -- & -- & -- & -- & -- & -- & -- & -- & -- & -- & -- & -- & -- & -- & 37.8 \\
 Liu eet al. \cite{LSWCCA20} & 34.5 & 37.1 & 33.6 & 34.2 & 32.9 & 37.1 & 39.6 & 35.8 & 40.7 & 41.4 & 33.0 & 33.8 &  33.0 & 26.6 & 26.9 & 34.7 \\
  Zeng et al. \cite{ZSHLXL20}& 34.8 & 32.1 & 28.5 & 30.7 & 31.4 & 36.9 & 35.6 & 30.5 & 38.9 & 40.5 & 32.5 & 31.0 & 29.9 & 22.5 & 24.5 & 32.0 \\
  Chen et al. \cite{CFSZCL21} & -- & -- & -- & -- & -- & -- & -- & -- & -- & -- & -- & -- & -- & -- & -- & 32.3 \\
  Zheng et al. \cite{ZZMYCD21}  & 30.0 & 33.6 & 29.9 & 31.0 & 30.2 & \textcolor{blue}{33.3} & 34.8 & 31.4 & 37.8 & 38.6 & 31.7 & 31.5 & 29.0 & 23.3 & 23.1 & 31.3 \\
  Li et al.  \cite{li2022mhformer} (T=351) & \textcolor{red}{27.7} & \textcolor{blue}{32.1} & 29.1 & 28.9 & 30.0 & 33.9 & \textcolor{blue}{33.0} & 31.2 & \textcolor{blue}{37.0} & 39.3 & 30.0 & 31.0 & 29.4 & 22.2 & 23.0 & 30.5 \\
 \hline
 \textbf{ConvFormer} (T=143) &  29.1 & 32.4 & \textcolor{blue}{28.1} & \textcolor{blue}{28.5} & \textcolor{red}{29.3} & \textcolor{blue}{33.3} & 33.3 & \textcolor{blue}{30.5} & \textcolor{blue}{37.0} & \textcolor{blue}{37.6} & \textcolor{red}{29.2} & \textcolor{red}{29.5} & \textcolor{blue}{28.4} & \textcolor{blue}{21.8} & \textcolor{red}{21.3} & \textcolor{blue}{29.9} \\
 \textbf{ConvFormer} (T=243) & \textcolor{blue}{28.9} & \textcolor{red}{31.8} & \textcolor{red}{28.0} & \textcolor{red}{28.2} & \textcolor{blue}{29.5} & \textcolor{red}{33.0} & \textcolor{red}{32.9} & \textcolor{red}{30.1} & \textcolor{red}{36.8} & \textcolor{red}{37.4} & \textcolor{blue}{29.8} & \textcolor{blue}{29.6} & \textcolor{red}{28.2} & \textcolor{red}{21.7} & \textcolor{blue}{21.5} & \textcolor{red}{29.8} \\
 
  \hline
  
    \hline
  \textbf{CPN-MPJPE (mm)} & \textbf{Dir.} & \textbf{Disc.} & \textbf{Eat} & \textbf{Greet} & \textbf{Phone} & \textbf{Photo} & \textbf{Pose}  & \textbf{Purch.} & \textbf{Sit} & \textbf{SitD.} & \textbf{Smoke} & \textbf{Wait} & \textbf{WalkD.} & \textbf{Walk}  & \textbf{WalkT.} & \textbf{Avg.} \\

 \hline
  Dabral et al. \cite{DMKASJ18} & 44.8 & 50.4 & 44.7 & 49.0 & 52.9 & 61.4 & 43.5 &  45.5 &  63.1 & 87.3 &  51.7 & 48.5 & 52.2 & 37.6 & 41.9 & 52.1 \\
  Cai et al.  \cite{CGLCCYT19} (T=7) & 44.6 & 47.4 & 45.6 & 48.8 & 50.8 & 59.0 & 47.2 & 43.9 & 57.9 & 61.9 & 49.7 &46.6 & 51.3 & 37.1 & 39.4 & 48.8 \\
  Pavllo et al. \cite{PFGA19} (T=243) & 45.2 & 46.7 & 43.3 & 45.6 & 48.1 & 55.1 & 44.6 & 44.3 & 57.3 & 65.8 & 47.1 & 44.0 &49.0 & 32.8 & 33.9 & 46.8 \\
 Lin and Lee \cite{LL19} (T=50) & 42.5 & 44.8 & 42.6 & 44.2 & 48.5 & 57.1 & 52.6 & 41.4 & 56.5 & 64.5 & 47.4 & 43.0 & 48.1 & 33.0 & 35.1 & 46.6 \\
 Yeh et al.  \cite{YHS19} & 44.8 & 46.1 & 43.3 & 46.4 & 49.0 & 55.2 & 44.6 & 44.0 & 58.3 & 62.7 & 47.1 & 43.9 & 48.6 & 32.7 & 33.3 & 46.7 \\
  Liu et al. \cite{LSWCCA20}  (T=243) & 41.8 & 44.8 & 41.1 & 44.9 & 47.4 & 54.1 & 43.4 & 42.2 & 56.2 & 63.6 & 45.3 & 43.5 & 45.3 & 31.3 & 32.2 & 45.1 \\
   Zeng et al. \cite{ZSHLXL20} & 46.6 & 47.1 & 43.9 & \textcolor{blue}{41.6} & 45.8 & 49.6 & 46.5 & \textcolor{red}{40.0} & 53.4 & 61.1 & 46.1 & 42.6 & 43.1 & 31.5 & 32.6 & 44.8 \\
   Wang et al.   \cite{WYXL20} (T=96) & 41.3 & 43.9 & 44.0 & 42.2 & 48.0 & 57.1 & 42.2 & 43.2 & 57.3 & 61.3 & 47.0 & 43.5 & 47.0 & 32.6 & 31.8 & 45.6 \\
   Chen et al. \cite{CFSZCL21} (T=243) & 41.4 & \textcolor{blue}{43.2} & 40.1 & 42.9 & 46.6 & 51.9 & 41.7 & 42.3 & 53.9 & \textcolor{red}{60.2} & 45.4 & 41.7 & 46.0 & 31.5 & 32.7 & 44.1 \\
    Lin et al. \cite{LWL21} (T=1) & -- & -- & -- & -- & -- & -- & -- & -- & -- & -- & -- & -- & -- & -- & -- & 54.0 \\
   Zheng et al. \cite{ZZMYCD21} (T=81) & 41.5 & 44.8 & 39.8 & 42.5 & 46.5 & \textcolor{blue}{51.6} & 42.1 & 42.0 & 53.3 & 60.7 & 45.5 & 43.3 & 46.1 &  31.8 & 32.2 & 44.3 \\
   Li et al. \cite{li2022mhformer} (T=351) & \textcolor{red}{39.2} & \textcolor{red}{43.1} & 40.1 & \textcolor{red}{40.9} & \textcolor{blue}{44.9} & \textcolor{red}{51.2} & \textcolor{red}{40.6} & 41.3 & 53.5 & \textcolor{blue}{60.3} & \textcolor{red}{43.7} & \textcolor{red}{41.1} & \textcolor{blue}{43.8} & 29.8 & \textcolor{red}{30.6} & \textcolor{red}{43.0} \\
 
 \hline
 \textbf{ConvFormer} (T=143) & 41.8 & 43.6 & \textcolor{blue}{39.3} & 43.2 & \textcolor{blue}{44.9} & 52.8 & 42.7 & 41.2 & \textcolor{blue}{53.1} & 60.9 & 45.0 & 41.9 & 44.7 & \textcolor{blue}{29.7} & 31.1 & 43.7 \\

 \textbf{ConvFormer} (T=243) & \textcolor{blue}{41.0} & \textcolor{blue}{43.2} & \textcolor{red}{39.0} & 42.4 & \textcolor{red}{44.5} & 52.2 & \textcolor{blue}{41.7} & \textcolor{blue}{40.8} & \textcolor{red}{53.0} & 60.6 & \textcolor{blue}{44.8} & \textcolor{blue}{41.3} & \textcolor{red}{43.7} & \textcolor{red}{29.6} & \textcolor{blue}{30.9} & \textcolor{blue}{43.2} \\
\hline

  \textbf{CPN-P-MPJPE (mm)} &\textbf{Dir.} & \textbf{Disc.} & \textbf{Eat} & \textbf{Greet} & \textbf{Phone} & \textbf{Photo} & \textbf{Pose}  & \textbf{Purch.} & \textbf{Sit} & \textbf{SitD.} & \textbf{Smoke} & \textbf{Wait} & \textbf{WalkD.} & \textbf{Walk}  & \textbf{WalkT.} & \textbf{Avg.} \\

 \hline
  Pavlakos et al. \cite{PZDD17} & 34.7 & 39.8 & 41.8 & 38.6 & 42.5 & 47.5 & 38.0 & 36.6 & 50.7 &  56.8 & 42.6 &39.6 & 43.9 & 32.1 & 36.5 & 41.5 \\
   Rayat et al. \cite{RL18} & 35.7 & 39.3 & 44.6 &43.0 & 47.2 & 54.0 & 38.3 & 37.5 & 51.6 & 61.3 & 46.5 & 41.4 &  47.3 & 34.2 & 39.4 & 44.1 \\
    Cai et al. \cite{CGLCCYT19} (T=7) & 35.7 & 37.8 & 36.9 & 40.7 & 39.6 & 45.2 & 37.4 & 34.5 & 46.9 & 50.1 & 40.5 & 36.1 & 41.0 & 29.6 &  32.3 & 39.0 \\
   Lin and Lee \cite{LL19} (T=50) & 32.5 & 35.3 & 34.3 &  36.2 & 37.8 & 43.0 & 33.0 & 32.2& 45.7 &  51.8 & 38.4 & 32.8 & 37.5 & 25.8 & 28.9 & 36.8 \\
  Pavllo et al. \cite{PFGA19} (T=243) & 34.1 & 36.1 & 34.4 & 37.2 & 36.4 & 42.2 & 34.4 & 33.6 & 45.0 & 52.5 & 37.4 & 33.8  & 37.8 & 25.6 & 27.3 & 36.5 \\
 Liu et al. \cite{LSWCCA20} (T=243) & 32.3 & 35.2 & 35.6 & \textcolor{blue}{34.4} & 36.4 & 42.7 & \textcolor{red}{31.2} & 32.5 & 45.6 & 50.2 & 37.3 & 32.8 & 36.3 & 26.0& 23.9 & 35.5\\
  Wang et al. \cite{WYXL20} (T=96) & 32.9 & 35.2 & 35.6 & \textcolor{blue}{34.4} & 36.4 & 42.7 & \textcolor{red}{31.2} & 32.5 & 45.6 & 50.2 & 37.3 & 32.8 & 36.3 & 26.0 & 23.9 & 35.5 \\
   Chen et al. \cite{CFSZCL21} (T=243) &  32.6 & 35.1 & 32.8  & 35.4 & 36.3 & 40.4 & 32.4 & 32.3 & \textcolor{red}{42.7} & 49.0 & 36.8 & 32.4 & 36.0 & 24.9 & 26.5 &  35.0 \\
   Zheng et al. \cite{ZZMYCD21} (T=81) & 32.5 & 34.8 & 32.6 & 34.6 & 35.3 & \textcolor{red}{39.5} & 32.1 & 32.0 & \textcolor{blue}{42.8} & \textcolor{red}{48.5} & \textcolor{red}{34.8} & 32.4 & 35.3 & 24.5 & 26.0 & 34.6 \\
   Li et al. \cite{li2022mhformer} (T=351) & \textcolor{blue}{31.5} & 34.9 & 32.8 & \textcolor{red}{33.6} & 35.3 & \textcolor{blue}{39.6} & \textcolor{blue}{32.0} & 32.2 & 43.5 & \textcolor{blue}{48.7} & 36.4 & 32.6 & \textcolor{red}{34.3} & 23.9 & \textcolor{blue}{25.1} & \textcolor{blue}{34.4} \\
 \hline
 \textbf{ConvFormer} (T=143) & 31.9 & \textcolor{blue}{34.4} & \textcolor{blue}{32.2} & 35.0 & \textcolor{blue}{34.2} & 40.7 & 32.9 & \textcolor{blue}{31.8} & 42.8 & 49.1 & \textcolor{blue}{36.0} & \textcolor{blue}{31.5} & 35.0 & \textcolor{blue}{23.6} & 25.2 & 34.5 \\
  \textbf{ConvFormer} (T=243) & \textcolor{red}{31.4} & \textcolor{red}{34.2} & \textcolor{red}{32.0} & 35.2 & \textcolor{red}{34.0} & 40.3 & 32.7 & \textcolor{red}{31.3} & \textcolor{red}{42.6} & 49.0 & 36.2 & \textcolor{red}{31.3} & \textcolor{blue}{34.8} & \textcolor{red}{23.4} & \textcolor{red}{24.9} & \textcolor{red}{34.2} \\
  \hline

\textbf{CPN-MPJVE} & \textbf{Dir.} & \textbf{Disc.} & \textbf{Eat} & \textbf{Greet} & \textbf{Phone} & \textbf{Photo} & \textbf{Pose}  & \textbf{Purch.} & \textbf{Sit} & \textbf{SitD.} & \textbf{Smoke} & \textbf{Wait} & \textbf{WalkD.} & \textbf{Walk}  & \textbf{WalkT.} & \textbf{Avg.} \\

\hline
 Pavllo et al. \cite{PFGA19} (T=243) & 3.0 & 3.1 & 2.2 & 3.4 & 2.3 &  2.7 &  2.7 &3.1 & 2.1 & 2.9 & 2.3 & 2.4 & 3.7 & 3.1 & 2.8 & 2.8 \\
  Chen et al. \cite{CFSZCL21} (T=243) & 2.7 &  2.8 & \textcolor{blue}{2.0} & 3.1 & \textcolor{blue}{2.0} & 2.4 & 2.4 &  2.8 & \textcolor{blue}{1.8} & \textcolor{blue}{2.4} & 2.0 & 2.1 & 3.4  & 2.7 & \textcolor{blue}{2.4} & 2.5 \\ 
  Wang et al. \cite{WYXL20} (T=96) & \textcolor{red}{2.3} & \textcolor{blue}{2.5} & \textcolor{blue}{2.0} & \textcolor{blue}{2.7} & \textcolor{blue}{2.0} & \textcolor{blue}{2.3} & \textcolor{blue}{2.2} & \textcolor{red}{2.5} & \textcolor{blue}{1.8} & 2.7 & \textcolor{blue}{1.9} & \textcolor{blue}{2.0} & \textcolor{blue}{3.1} & \textcolor{red}{2.2} & 2.5 & \textcolor{blue}{2.3} \\
\hline
\textbf{ConvFormer} (T=143) & \textcolor{red}{2.3} & \textcolor{red}{2.3} & \textcolor{red}{1.8} & \textcolor{red}{2.6} & \textcolor{red}{1.8} & \textcolor{red}{2.1} & \textcolor{red}{2.1} & \textcolor{red}{2.5} & \textcolor{red}{1.4} & \textcolor{red}{2.0} & \textcolor{red}{1.7} & \textcolor{red}{1.9} & \textcolor{red}{3.0} & \textcolor{blue}{2.4} & \textcolor{red}{2.1} & \textcolor{red}{2.1} \\
\hline
\textbf{GT-MPJVE} & \textbf{Dir.} & \textbf{Disc.} & \textbf{Eat} & \textbf{Greet} & \textbf{Phone} & \textbf{Photo} & \textbf{Pose}  & \textbf{Purch.} & \textbf{Sit} & \textbf{SitD.} & \textbf{Smoke} & \textbf{Wait} & \textbf{WalkD.} & \textbf{Walk}  & \textbf{WalkT.} & \textbf{Avg.} \\

\hline
 Wang et al. \cite{WYXL20} (T=96) & \textcolor{red}{1.2} & \textcolor{red}{1.3} & \textcolor{blue}{1.1} & \textcolor{red}{1.4} & \textcolor{blue}{1.1} & \textcolor{blue}{1.4} & \textcolor{red}{1.2} & \textcolor{red}{1.4} & \textcolor{blue}{1.0} & \textcolor{blue}{1.3} & \textcolor{blue}{1.0} & \textcolor{red}{1.1} & \textcolor{red}{1.7} & \textcolor{red}{1.3} & \textcolor{blue}{1.4} & \textcolor{blue}{1.4} \\
\hline
\textbf{ConvFormer} (T=143) &  \textcolor{red}{1.2} & \textcolor{red}{1.3} & \textcolor{red}{0.9} & \textcolor{red}{1.4} & \textcolor{red}{1.0} & \textcolor{red}{1.2} & \textcolor{blue}{1.3} & \textcolor{blue}{1.5} & \textcolor{red}{0.7} & \textcolor{red}{1.1} & \textcolor{red}{0.9} & \textcolor{red}{1.1} & \textcolor{red}{1.7} & \textcolor{blue}{1.4} & \textcolor{red}{1.2} & \textcolor{red}{1.2}\\
\hline
  
\end{tabular}}
\label{Table:GT}

\end{table*}

\begin{table*}[!htb]

\centering
\caption{\small Quantitative results on HumanEva under protocol 2 for the left part of the table and quantitative results on MPI-INF-3DHP in the right part of the table. Best is in \textcolor{red}{Red} and second best is \textcolor{blue}{Blue}.}

 \resizebox{\columnwidth}{!}{\begin{tabular}{*{14}{|c}|}

  \hline
  Action &  & Walk & & & Jog &  &  & Box & &  Method & PCK $\uparrow$ & AUC $\uparrow$ & MPJPE (mm) $\downarrow$ \\
  \hline
  Subject &  S1 & S2 & S3 &  S1 & S2 & S3  & S1 & S2 & S3 &   \cite{mono-3dhp2017} & 75.7 & 39.3 & 117.6 \\

 \hline
   Martinez et al. \cite{MHRL17} & 19.7 & 17.4 & 46.8 & 26.9 & 18.2 & 18.6 &  -- &  -- &  -- &   Lin et al. \cite{LL19} & 83.6 & 51.4 & 79.8 \\
    Pavalkos et aal. \cite{PZDD17} & 22.3 & 19.5 & 29.7 & 28.9 & 21.9 & 23.8 & -- & -- & -- &  Pavllo et al. \cite{PFGA19} & 86.0 & 51.9 & 84.0\\
  Pavllo et al. \cite{PFGA19} & 13.9 & 10.2 & 46.6 & 20.9 & 13.1 & 13.8 & 23.8 & 33.7 & 32.0 &  Li et al. \cite{LKPTTC20}  & 81.2 & 46.1 & 99.7 \\
  Zheng et al. \cite{ZZMYCD21} & 16.3 & 11.0 & 47.1 & 25.0 & 15.2 & 15.1 & -- & -- & -- &    Chen et al. \cite{CFSZCL21} & 87.6 & 54.0 & 78.8 \\
 \hline
 \textbf{ConvFormer} (T=9) & 12.5 & 10.1 & 25.4 & 13.3 & 12.9 & 22.6 & 31.7 & 28.6 & 29.0 &    Zheng et al. \cite{ZZMYCD21} & 88.6 & 56.4 & 77.1 \\
 \textbf{ConvFormer} (T=27) & \textcolor{blue}{11.4} & \textcolor{blue}{9.0} & \textcolor{blue}{20.1} & \textcolor{blue}{19.1} & \textcolor{blue}{11.8} & \textcolor{blue}{11.8} & \textcolor{blue}{20.8} & \textcolor{blue}{28.0} & \textcolor{blue}{26.1} &  Li et al.  \cite{li2022mhformer} & \textcolor{blue}{93.8} & \textcolor{blue}{63.3} & \textcolor{blue}{58.0} \\
 \textbf{ConvFormer} (T=43) & \textcolor{red}{10.7} & \textcolor{red}{7.9} & \textcolor{red}{16.0} & \textcolor{red}{16.7} & \textcolor{red}{9.3} & \textcolor{red}{10.0} & \textcolor{red}{18.2}
 & \textcolor{red}{25.0} & \textcolor{red}{24.3} &  \textbf{ConvFormer} & \textcolor{red}{96.4} & \textcolor{red}{69.8} & \textcolor{red}{53.6} \\

\hline
\end{tabular}}

\label{Table:HumanEva}
\end{table*}

We report results for our $143$ and $243$ frame models on H3.6M and we report results for our $9$, $27$, and $43$ frame model for HumanEva. We report all 15 action results for both subjects S9 and S11 using GT and CPN detections as the 2D input under protocol I, II, and III in Table. \ref{Table:GT} and the last column represents the average. \textbf{ConvFormer's $\textbf{143}$ and $\textbf{243}$-frame models substantially reduce the parameter count by $\textbf{83.4}$\textbf{\%} and $\textbf{65.5}$\textbf{\%} respectively, relative the previous SOTA} \cite{li2022mhformer}. ConvFormer's $143$ and $243$-frame models outperforms the previous SOTA on GT inputs -- achieving a $2.3$\% reduction of error. ConvFormer's $243$-frame model misses SOTA on CPN inputs for Protocol I by $0.2$mm while having substantially lowered parameters and achieving best or second best on 11 of the 15 actions. However, it outperforms the SOTA on some challenging actions such as \textit{Sitting} and \textit{WalkingDog} which exhibit complex postures and rapid postural changes.   Under Protocol II ConvFormer achieves SOTA on 9 individual actions and on the average error. Lastly, for both GT and CPN inputs ConvFormer reduces the MPJVE by $8.6$\% and $14.3$\% respectively, resulting in smoother predictions. See Figure \ref{fig:H36MExamples} for some qualitative results on H36M or see \url{https://github.com/AJDA1992/ConvFormer} for more examples from challenging in-the-wild motions.

The left side of Table \ref{Table:HumanEva} shows the results of training ConvFormer from scratch on HumanEva. We note that our larger receptive field model, with 43 frames, achieves SOTA for every action, while our  27 frame receptive field model achieves second place for every action.

The right side of Table \ref{Table:HumanEva} reports the quantitative results of ConvFormer on MPI-INF-3DHP relative to other methods. Following \cite{ZZMYCD21,li2022mhformer}, we use 2D pose sequences of 9 frames due to fewer samples and shorter video sequences. We note that ConvFormer increases PCK by $2.7$\%, AUC by $10.2$\%, and decreases MPJPE by $7.6$\%.

\subsection{Ablations}
First, we study the contribution of individual hyper-parameters and tune them. Second, we assess the contribution of convolutional self-attention relative to the baseline (vanilla transformer) and then the contribution of our dynamic self-attention mechanism. To the first point, we perform an extensive grid search using \cite{liaw2018tune} and report some of the results in Tables \ref{tab:DimensionsBlocks}, \ref{tab:kernels}. We fine the following hyper-parameters to be optimal: $d=32$, $B_{sp}=B_{temp}=2$ and using the following kernel sizes $(7,7,7)$.

In Table \ref{tab:sequence_length} we analyze the effect of receptive field alongside parameter counts relative to other transformer based methods. We fix the optimal hyper-parameters found in Tables \ref{tab:DimensionsBlocks}, \ref{tab:kernels}. We find across all receptive fields, ConvFormer reduces parameters substantially relative \cite{li2022exploiting,ZZMYCD21} while remaining extremely competitive on CPN inputs for Protocol I. 

Finally, we analyze what improvement ConvFormer brings relative to a vanilla transformer architecture and the benefit of you using our Dynamic Multi-Headed attention mechanism. In Table \ref{tab:ablation} our baseline model following the same architecture as ConvFormer except with class scaled dot product attention and fully-connected layers generating the queries, keys, and values. We find by using a single filter in our ConvFormer architecture improves on the baseline by 2mm and introducing our Dynamic Multi-Headed Attention we reduce by another 1.1 mm.

\begin{table*}[!htb]
    \centering
    \caption{\small Analysis of spatial embedding dimension and number of spatial and temporal ConvFormer Blocks. We also perform a limited analysis on number of attention heads. Optimal performance is marked by \textcolor{red}{Red} and assessed by MPJPE.}
 \resizebox{\columnwidth}{!}{\begin{tabular}{ *{14}{|c}|}
    \hline
    d & 16 & 16 & 16 & \textcolor{red}{32} & 32 & 32 & 64 & 32 & 32 & 32 \\
    \hline
    $B_{sp}$  &  2 &  2 & 4 & \textcolor{red}{2} & 2 & 4 & 2 & 2 & 2 & 2 \\
    \hline
    $B_{temp}$ &  2 & 4 & 2 & \textcolor{red}{2} & 4 & 2 & 2 & 2 & 2 & 2 \\
    \hline
    Params (M) & 0.65 & 1.26 & 0.69 & \textcolor{red}{2.56} & 4.95 & 2.70 & 9.97 & 2.56 & 2.56 & 2.56 \\
    \hline
    Heads & 8 & 8 & 8 & \textcolor{red}{8} & 8 & 8 & 8 & 1 & 2 & 4 \\
    \hline
    MPJPE (mm) & 50.8 & 51.0& 51.7 & \textcolor{red}{49.4} & 50.6 & 49.9 & 50.1 & 52.3 & 51.0 & 49.6 \\
    \hline
    \end{tabular}}
    \label{tab:DimensionsBlocks}

\end{table*}

\begin{table}[!htb]

    \centering
    \caption{\small Analysis of different kernel configurations where performance is evaluated relative to MPJPE on CPN detections for H36M. Best marked in \textcolor{red}{Red}. }
    \resizebox{.6\columnwidth}{!}{\begin{tabular}{ *{6}{|c}|} 
    \hline
     $B_{sp}$ & $B_{temp}$ & Kernels & MPJPE (mm) & Params (M)   \\
     \hline

       2 & 2 & 3 & 51.7 & 2.44 \\
       2 & 2 & 3,3 & 51.6 & 2.46   \\
       2 & 2 & 3,3,3 & 50.8 & 2.48 \\
       2 & 2 & 5 & 50.5 & 2.46  \\
       2 &  2 & 5,5 & 52.0 & 2.49 \\
       2 & 2 & 5,5,5 & 51.9 & 2.52 \\
       2 & 2 & 7 & 50.5 & 2.47 \\
       2 & 2 & 7,7 & 50.1 & 2.51 \\
       \textcolor{red}{2} & \textcolor{red}{2} &  \textcolor{red}{7,7,7} & \textcolor{red}{49.4}  & \textcolor{red}{2.56}\\
       2 & 2 & 9 & 51.5 & 2.48 \\
       2  & 2 & 9,9 & 50.8 & 2.54\\
     
       2 & 2 & 9,9,9 & 50.6 & 2.60 \\
      2 & 2 & 3,5,7 & 50.6 & 2.52 \\
      2 & 2 & 5,7,9 & 51.6 & 2.56 \\

     \hline
     \end{tabular}}
    \label{tab:kernels}

\end{table}

\begin{table}[!htb]

    \centering
    \caption{\small Parameter count and FLOPs results with MPJPE for different transformer architectures and graph attention networks separated by receptive field. The last grouping is for models with largest receptive field. Best and second best marked with \textcolor{red}{Red} and \textcolor{blue}{Blue} respectively.}
 
    \resizebox{.6\columnwidth}{!}{\begin{tabular}{ *{5}{|c}|} 
    \hline
     & T & Params (M)  &  FLOPs (M)* & MPJPE (mm)  \\
     \hline

    Zheng et al. \cite{ZZMYCD21} & 9 & \textcolor{blue}{9.58} & \textcolor{blue}{180} & 49.9 \\
     Li et al. \cite{li2022mhformer} & 9 & 19.09  & 340 & \textcolor{red}{47.8}   \\
     \textbf{ConvFormer} & 9 & \textcolor{red}{2.56} & \textcolor{red}{100} & \textcolor{blue}{49.4}   \\
     \hline
     Zheng et al. \cite{ZZMYCD21} & 27  & \textcolor{blue}{9.59} &  \textcolor{blue}{540} & \textcolor{blue}{47.0}   \\
     Li et al. \cite{li2022mhformer} & 27 & 19.18 & 1040 & \textcolor{red}{45.9} \\ 
     \textbf{ConvFormer} & 27 & \textcolor{red}{2.65} & \textcolor{red}{360} & 47.7   \\
     \hline
     Zheng et al. \cite{ZZMYCD21} & 81 &  \textcolor{blue}{9.60} & \textcolor{blue}{1620} & \textcolor{red}{44.3}   \\
     Li et al. \cite{li2022mhformer} & 81 & 19.84 & 3120 & \textcolor{blue}{44.5}  \\
      \textbf{ConvFormer} & 81 & \textcolor{red}{3.43} & \textcolor{red}{1600}  & 45.0   \\
      \hline
      Liu et al. \cite{Lgraph21} & 243 & \textcolor{blue}{7.09} & \textcolor{blue}{9700} & 44.9 \\
     Li et al. \cite{li2022mhformer} & 351 & 31.52 &  14160 & \textcolor{red}{43.0}  \\
     \textbf{ConvFormer} & 143 & \textcolor{red}{5.24} & \textcolor{red}{4220} & 43.7   \\
     \textbf{ConvFormer} & 243 & 10.24 & 10000 & \textcolor{blue}{43.2} \\
     \hline
     \end{tabular}}

    \label{tab:sequence_length}

\end{table}

\begin{table}[]
    \centering
    \begin{tabular}{|c|c|c|}
    \hline
        \textbf{Method} & Params (M) & MPJPE (mm) \\
        \hline
         Baseline & 5.2 & 52.4 \\
         Single-Filter ConvFormer & 2.47 & 50.5 \\
         Dynamic ConvFormer & 2.56 &  49.4 \\
         \hline
    \end{tabular}
    \caption{\small Ablation study analyzing effect of different components of ConvFormer on Accuracy and Parameter Counts.}
    \label{tab:ablation}
\end{table}

\section{Conclusion}
In this paper we attempt to address the ever growing complexity of transformer models. For this, we introduce ConvFormer which is based on three novel components: temporal fusion, convolutional self-attention, and dynamic feature aggregation. To assess the effectiveness of different components we conducted extensive ablation studies. We reduced the parameter counts relative to the previous SOTA by over $\textbf{65}$\textbf{\%} while achieving SOTA on H36M for Protocol I on GT inputs, Protocol II for CPN detections, Protocol III for both GT and CPN inputs, HumanEva for all subjects, and lastly all three metrics of MPI. Interestingly, even though graph convolutional networks and graph attention networks are light-weight and robustly model spatial/temporal relationships, ConvFormer provides a better trade off between error reduction and computational complexity. We believe ConvFormer will provide more ready access to high quality 3D reconstruction networks by making the training and inference process less computationally demanding.

\clearpage

\section{Appendices}
\subsection{Attention Visualization}

\begin{figure}[!htb]
    \centering
    \includegraphics[width=\textwidth]{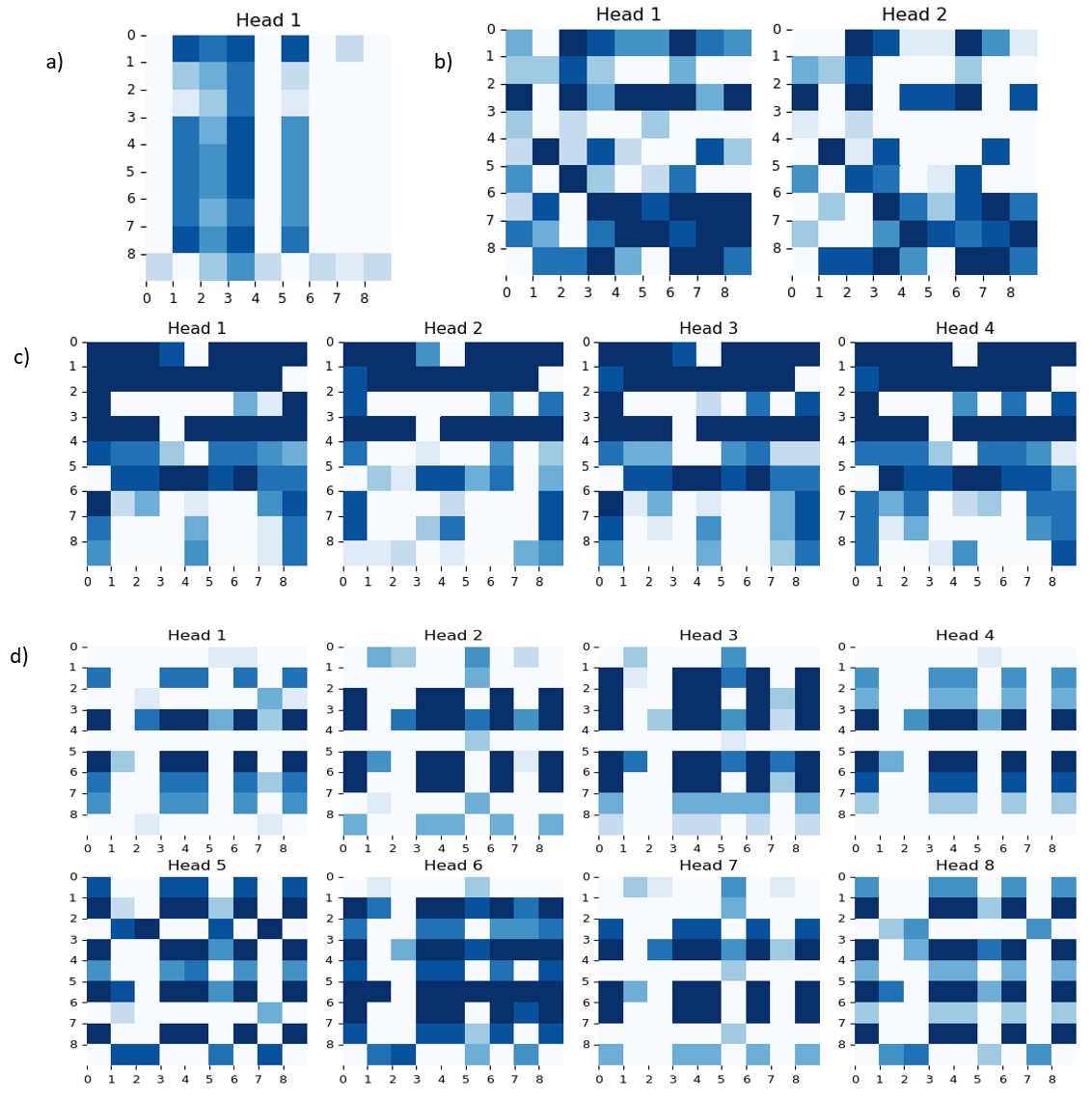}
    \caption{\small Temporal attention for 9 frame ConvFormer trained on H36M using CPN detections as input -- a) is our one head model, b) is 2 head model, c) is 4 head and d) is the 8 head model which achieves lowest MPJPE.}
    \label{fig:attentionablation}
\end{figure}

We provide in figure \ref{fig:attentionablation} additional visualizations of temporal attention heads for part of our ablation on number of attention heads with quantitative results reported in Table \ref{tab:DimensionsBlocks}. We note that our $8$ head model achieves the lowest MPJPE on H3.6M. We hypothesize that, although there is a clear visual indication that as the number of heads increases, redundancies within the attention maps occur with subtle variations, that this redundancy acts as a noise filtering mechanism by highlighting critical information. In the NLP landscape an extensive analysis on BERT was conducted to understand the optimal number of heads and how one can perform head pruning during test-time without substantial performance impact, \cite{MLN19}.

\begin{figure}[!htb]
    \centering
    \subfigure{\includegraphics[width=\textwidth]{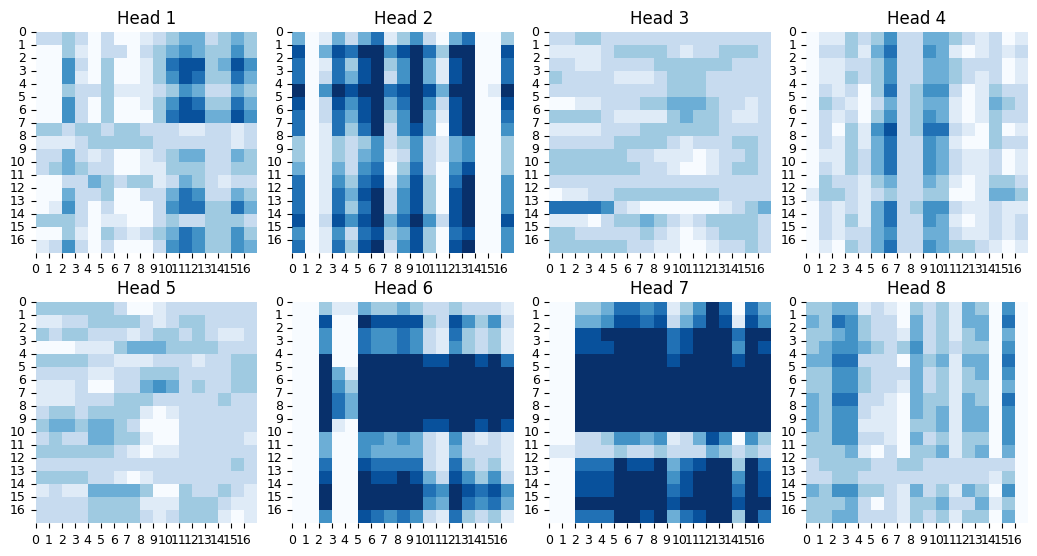}} 
    \subfigure{\includegraphics[width=\textwidth]{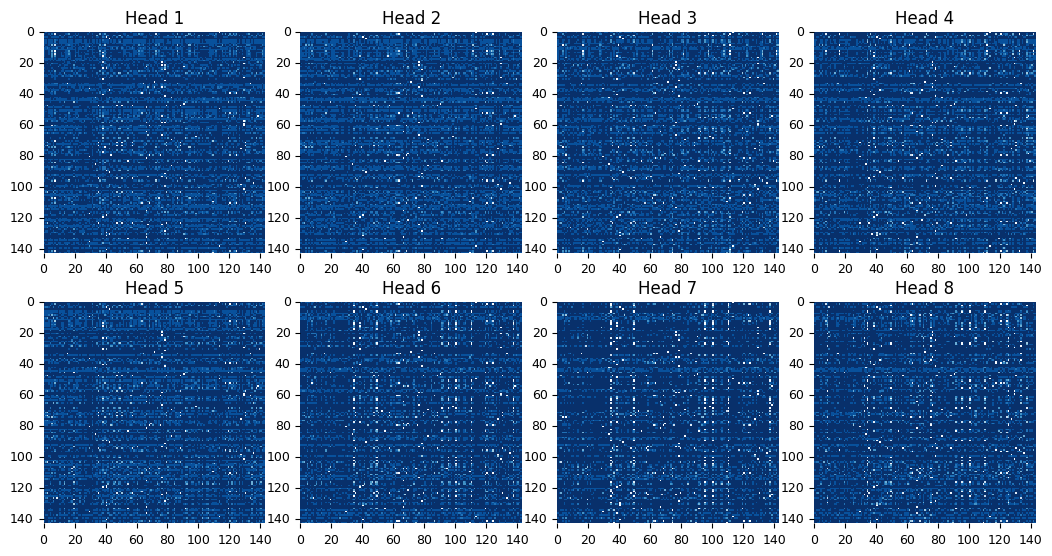}} 
    \caption{\small Example of Attention maps, top is the spatial ConvFormer and the bottom is the temporal ConvFormer for the 143 frame model trained on CPN detections for H3.6M. These maps were generated for S9 for the Directions action.}
    \label{fig:attention}
\end{figure}

We also provide visualizations of the attention heads for both the spatial and temporal ConvFormer for all attention  heads utilized in our model. We evaluate the attention heads for subject 9 from H3.6M for the \textit{Directions} action. The spatial self attention maps are seen in Figure \ref{fig:attention} and the x-axis corresponds to the 17 joints of the H3.6M skeleton and the y-axis corresponds to the attention output. These maps correspond to the 143 frame model and the temporal attention maps x-axis is the 143 frames of the sequence while the y-axis is the attention at each frame. The attention heads return different attention magnitudes which represent either spatial correlations or frame-wise global correlations learned from the joint temporal profiles. 

\clearpage

\clearpage
\begin{thebibliography}{9}

\bibitem{MHRL17} J. Martinez, R. Hossain, J. Romero, aand J. J. Little, 
  title={A simple yet effective baseline for 3d human pose estimation}, A simple yet effective baseline for 3d human pose estimation, Proceedings of the IEEE International Conference on Computer Vision, (2017), 2640-2649.


\bibitem{PFGA19} D. Pavllo, C. Feichtenhofer, D. Grangier, and M. Auli, 3d human pose estimation in video with temporal convolutions and semi-supervised training, Proceedings of the IEEE Conference on Computer Vision and Pattern Recognition (CVPR), (2019), 7753-7762.

\bibitem{liaw2018tune}R. Liaw, E. Liang, R. Nishihara, P. Moritz, J. Gonzalez, and I. Stoica, Tune: A Research Platform for Distributed Model Selection and Training, arXiv preprint arXiv:1807.05118, (2018).

\bibitem{KTSLSF14} A. Karpathy, G. Toderici, S. Shetty, T. Leung, R. Sukthankar, and L. Fei-Fei, Large-Scale Video Classification with Convolutional Neural Networks, Proceedings of the IEEE Conference on Computer Vision and Pattern Recognition (CVPR), (2014), 1725-1732.

\bibitem{ZZMYCD21} C. Zheng, S. Zhu, M. Mendieta, T. Yang, C. Cheng, and Z. Ding, 3D Human Pose Estimation with Spatial and  Temporal Transformers, Proceedings of the IEEE International Conference on Computer Vision (ICCV), (2021).

\bibitem{FXWLZ18} H. S. Faang, Y. Xu, W. Wang, X. Liu, and S. C. Zhu, Learning pose grammar to encode human body configuration for 3d pose estimation, Thirty-Second AAAI Conference on Artificial Intelligence, (2018).

\bibitem{IPOS13} C. Ionescu, D. Papava, V. Olarue, C. Sminchisescu, Human3.6m: Large scale datasets and predictive methods for 3d human sensing in natural environments, IEEE transactions on pattern analysis and machine intelligence, 36 (7), (2013), 1325-1339.

\bibitem{SBB10} L. Sigal, A. O. Balaan, and M. J. Black, Humaneva: Synchronized video and motion capture dataset and baseline algorithm for evaluation of articulated human motion, IEEE Transactions on Pattern Analysis and Machine Intelligence, 87 (12), (2010), 4-27.

\bibitem{PZDD17} G. Pavaalkos, X. Zhou, K. G. Derpanis, and K. Daniilidis, Coarse-to-fine volumetric prediction for single-image 3d human pose, Proceedings  of the  IEEE Conference on Computer Vision and Pattern Recognition (CVPR), (2017), 7025-7034.

\bibitem{mono-3dhp2017} D. Mehta, H. Rhodin, D. Casas, P. Fua, O. Sotnychenko, W. Xu, and C. Theobalt, Monocular 3D Human Pose Estimation In The Wild Using Improved CNN Supervision,3D Vision (3DV), 2017 Fifth International Conference on (2017


\bibitem{CFSZCL21}, T. Chen, C. Fang, X. Shen, Y. Zhu, Z. Chen, and J. Luo, Anatomy-aware 3d human pose estimation with bone-based pose decomposition, IEEE Transactions on Circuits and Systems for Video Technology, (2021).

\bibitem{VSPUJGKP17} A. Vaswani, N. Shazeer, N. Parmar, J. Uszkoreit, L. Jones, A. N. Gomez, L. U. Kaiser, and I. Polosukhin, Attention is all you need, Advances in Neural Information Processing Systems (NIPS), (2017), 5998-6008.

\bibitem{LSWCCA20} R. Liu, J. Shen, H. Wang, C. Chen, S. C. Cheung, and V. Asari, Attention mechanism exploits temporal contexts: Real-time 3d human pose reconstruction, Proceedings of the IEEE Conference on Computer Vision and Pattern Recognition (CPRV), (2020), 5064-5073.

\bibitem{KKA19} M. Kocabas, S. Karagozz, and E. Akbas, Self-supervised learning of 3d human pose using multi-view geometry, Proceedings of the IEEE Conference on Computer Vision and Pattern Recognition (CVPR), (2019), 1077-1086.

\bibitem{RL18} M. Raayat Imtia Hossain and J. J. Little, Exploiting temporal information for 3d human pose estimation, Proceedings of the European Conference on Computer Vision (ECCV), (2018), 68-84.

\bibitem{WYXL20} J. Wang, S. Yan, Y. Xiong, and D. Lin, Motion guided 3d pose estimation from videos, ArXiv, arXiv:2004.13985, (2020).

\bibitem{PGCCYDLDAL17} A. Paszke, S. Gross, S. Chintala, G. Chanan, E. Yang, Z. DeVito, Z. Lin, A. Desmaison, L. Antiga, and A. Lerer, Automatic Differentiation in PyTorch, NIPS 2017 Workshop on Autodiff, (2017).

\bibitem{HSLSW16} J. Wang, S. Yan, Y. Xiong, and D. Lin, Deep networks with stochastic depth, European Conference on Computer Vision (ECCV), (2016), 646-661.

\bibitem{CH-MSWS19} Z. Cao, G. Hidalgo, T. Simon, S-E. Wei, and Y. Sheikh, OpenPose: realtime multi-person 2D pose estimation using Part Affinity Fields, IEEE transactions on pattern analysis and machine intelligence, 43 (1), (3019), 172-186.

\bibitem{WRKS16} S-E. Wei, V. Ramakrishna, T. Kanade, and Y. Sheikh, Convolutional pose machines, Proceedings of the IEEE conference on Computer Vision and Pattern Recognition, (2016), 4724-4732.

\bibitem{KDWHUBMDHGUZ21} A. Kolesnikov, A. Dosovitskiy, D. Weissenborn, G. Heigold, J. Uszkoreit, L. Beyer, M. Minderer, M. Dehghani, N. Houlsby, S. Gelly, T. Unterthiner, and X. Zhai, An Image is Worth 16x16 Words: Transformers for Image Recognition at Scale, (2021).

\bibitem{JMLR:v15:srivastava14a} N. Srivastava, G. Hinton, A. Krihevsky, I. Sutskever, and R. Salakhudinov, Dropout: A Simple Way to Prevent Neural Networks from Overfitting, Journal of Machine Learning Research, 15 (56), (2014), 1929-1958.

\bibitem{FXTL17} H-S. Fang, S. Xie, Y-W. Tai, and C. Lu, RMPE: Regional Multi-person Pose Estimation, International Conference on Computer Vision (ICCV), (2017).
                  
\bibitem{CWPZY18} Y. Chen, Z. Wang, X. Peng, Z. Zhang, G. Yu, and J. Sun, Cascaded Pyramid Network for Multi-person Pose Estimation, IEEE/CVF Conference on Computer Vision and Pattern Recognition (CVPR), (2018), 7103-7112.

\bibitem{SXLW19} K. Sun, B. Xiao, D. Liu, and J. Wang, Deep High-Resolution Representation Learning for Human Pose Estimation, IEEE/CVF Conference on Computer Vision and Pattern Recognition (CVPR), (2019).

\bibitem{XWW18} B. Xiao, H. Wu, Y. Wei, Simple Baselines for Human Pose Estimation and Tracking, European Conference on Computer Vision (ECCV), (2020).

\bibitem{WLLLH20} Z. Wu, Z. Liu, J. Lin, and S. Han, Lite transformer with long-short range attention, International Conference on Learning Representations (ICLR), (2020).

\bibitem{JYZCH20} Z. Jiang, W. Yu, D. Zhou, Y. Chen, J. Feng, and S. Yan, Convbert: Improving bert with span-based dynamic convolution, Advances in Neural Information Processing Sytems (NeurIPS), (2020).


\bibitem{CDLCYL20} Y. Chen, X. Dai, M. Liu, D. Chen, L. Yuan, and Z. Liu, Dynamic Convolution: Attention Over Convolution Kernels, IEEE/CVF Conference on Computer Vision and Pattern Recognition (CVPR), (2020), 11027-11036.

\bibitem{ZSHLXL20} A. Zeng, X. Sun, F. Huaang, M. Liu, Q. Xu, and S. Lin, Srnet: Improving generalization in 3d human pose estimation with a split-and-recombine approach, European Conference on Computer Vision (ECCV), (2020).

\bibitem{DMKASJ18} R. Dabral, A. Mundhaada, U. Kusupati, S. Afaque, A. Sharma, and A. Jain, Learning 3D Human Pose from Structure and Motion, European Conference on Computer Vision (ECCV), (2018).

\bibitem{CGLCCYT19} Y. Cai, L. Ge, J. Liu, J. Cai, T-J. Cham, J. Yuan, N. M. Thalmann, Exploiting Spatial-Temporal Relationships for 3D Pose Estimation via Graph Convolutional Networks, 2019 IEEE/CVF International Conference on Computer Vision (ICCV), (2019), 2272-2281.

\bibitem{LL19} J. Lin and G. Lee, Trajectory Space Factorization for Deep Video-Based 3D Human Pose Estimation, British Machine Vision Conference (BMVC), (2019).

\bibitem{YHS19} R. Yeh, Y. T. Hu, and A. Schwing, Chirality Nets for Human Pose Regression, International Conference on Neural Information Processing Systems (NeurIPS), (2019).

\bibitem{LWL21} K. Lin, L. Wang, aand Z. Liu, End-to-End Human Pose and Mesh Reconstruction with Transformers, IEEE/CVF Conference on Computer Vision and Pattern Recognition (CVPR), (2021).

\bibitem{ZKLOT16} B. Zhou, A. Khosla, A. Lapedriza, A. Oliva, and A. Torralba, Learning Deep Features for Discriminative Localization, IEEE Conference on Computer Vision and Pattern Recognition (CVPR), (2016), 2921-2929.


\bibitem{MCL19} G. Moon, J. Chang, and K. M. Lee, Camera Distance-aware Top-down Approach for 3D Multi-person Pose Estimation from a Single RGB Image, The IEEE Conference on International Conference on Computer Vision (ICCV), (2019).

\bibitem{PZD18} G. Pavlakos, X. Zhou, and K. Daniilidis, Ordinal depth supervision for 3D human pose estimation, IEEE/CVF Conference on Computer Vision and Pattern Recognition (CVPR), (2018).

\bibitem{ML20} G. Moon and K. M. Lee, I2l-meshnet: Image-to-lixel prediction network for accurate 3d human pose estimation and mesh estimation from a single rgb image, European Conference on Computer Vision (ECCV), (2020).

\bibitem{KNHZKS21} S. Khan, M. Naseer, M. Hayat, S. Zamir, F. Khan, and M. Shah, Transformers in Vision: A surver, ArXiv, arXiv:2101.01169, (2021).

\bibitem{CMSUKZ20} N. Carion, F. Massa, G. Synnaeve, N. Usunier, A. Kirillob, and S. Zagoruyko, Ene-to-End Object Detection with Transformers, ArXiv, http://arxiv.org/abs/2005.12872, (2020).

\bibitem{LB14} S. Li and A. B. Chan, 3D Human Pose Estimation from Monocular Images with Deep Convolutional Neural Network, Asian Conference on Computer Vision (ACCV), (2014). 

\bibitem{LKPTTC20} S. Li, L. Ke, K. Pratama, Y. W. Tai, C. K. Tang, K. T. Cheng, Cascaded deep monocular 3D human pose  estimation with evolutionary training data, ArXiv arXiv:2006.07778 (2020).

\bibitem{SXWLW17} X. Sun, B. Xiao, S. Liang, Y. Wei, Integral human pose regression, ArXiv arXiv:1711.08229, (2017).

\bibitem{KBJM17} A. Kanmazawa, M. J. Black, D. W. Jacobs, and J. Malik, End-to-end Recovery of Human Shape and Pose, IEEE/CVF Conference on Computer Vision and Pattern Recognition (CVPR), (2017).

\bibitem{zhou2019hemlets} K. Zhou, X. Han, N. Jaing, K. Jia, and J. Lu, HEMlets Pose: Learning Part-Centric Heatmap Triplets for Accurate 3D Human Pose Estimation, International Conference of Computer Vision (ICCV), (@019).

\bibitem{CYWWT19} Y. Cheng, B. Yang, B. Waang, Y. Wending, and R. T. Tang, Occlusion Aware Networks for 3D Human Pose Estimation in Video, International Conference of Computer Vision (ICCV), (2019).

\bibitem{ZHJJL21} K. hou, X. Han, N. Jiang, K. Jai, aand J. Lu, HEMlets PoSh: Learning Part-Centric Heatmap Triplets for 3D Human Pose and Shape  Estimation, IEEE Transactions on Pattern Analysis and Machine Intelligence (TPAMI). (2021).

\bibitem{HZLLX19} F. Huang, A. Zeng, M. Liu, Q. Lai, and Q. Xu, DeepFuse: An IMU-Aware Network for Real-Time 3D Human Pose Estimation from Multi-View Image, IEEE Winter Conference on Applications of Computer Vision (WACV), (2020), 418-427.

\bibitem{SWL21} H. Shuai, L. Wu, and Q. Liu, Adaptively Multi-view and Temporal Fusing Transformer for 3D Human Pose Estimation, ArXiv abs/2110.05092, (2021).


\bibitem{IBLM19} K. Iskakov, E. Burkov, V. Lempitsky, and Y. Malkov, Learnable Triangulation of Human Pose. IEEE/CVF International Conference on Computer Vision (ICCV), (2019).

\bibitem{QWWWZ19} H. Qiu, C. Wang, J. Wang, N. Wang, and W. Zeng, Cross View Fusion for 3D Human Pose Estimation, IEEE/CVF International Conference on Computer Vision (ICCV), (2019), 4341-4359.

\bibitem{DJHBZ19} J. Dong, W. Jiang, Q. Huang, H. Bao, X. hou, Fast and Robust Multi-Person 3D Pose Estimation from Multiple View, IEEE/CVF Conference on Computer Vision and Pattern Recognition (CVPR), (2019).

\bibitem{YRKS20} Y. He, R. Yan, K. Fragkiadaki, S. Yu. Epipolr Transformers, IEEE/CVF Conference on Computer Vision and Pattern Recognition (CVPR), (2020), 7776-7785.


\bibitem{YCWYSJTFY21} L. Yuan, Y. Chen, T. Wang, W. Yu, Y. Shi, . Jiang, F. Tay, J. Feng, and S. Yan, Tokens-to-Token ViT: Training Vision Transformers from Scratch on ImageNet, IEEE International Conference on Computer Vision (ICCV) (2021), 538-547

\bibitem{ZSWJYCL20} Z. Liu, N. Shun, W. Li, J. Lu, Y. Wu, C. Li, aand L. Yang, ConvTransformer: A Convolutional Transformer Network for Video Frame Synthesis, ArXiv, abs/2011.10185 (2020).

\bibitem{HMMFA21} H. Touvron, M. Cord, M. Doue, F. Massa, A. Sablayrolles, H. J'egou, Training data-efficient image transformers and distillation through attention, International Conference of Machine Learning (ICML), (2021).

\bibitem{Sharir2021AnII} S. Gilad, N. Asaf, -M. Lihi, An Image is Worth 16x16 Words, What is a Video Worth?, ArXXiv, abs/2103.13915 (2021).

\bibitem{CJHR21} K. Chaitanya, M. Joshua, D. Hang, M-S. Roderick, CpT: Convolutional Point Transformer for 3D Point Cloud Processing, ArXXiv, abs/2111.10866 (2021).

\bibitem{PA21} P. Paaschalis and A. Antonis, PE-former: Pose Estimation Transformer, CoRR abs/2112.04981, (2021).

\bibitem{kitaev2020reformer} N. Kitaev, L. Kaiser, and A. Levskaya, Reformer: The Efficient Transformer, Proceedings of International Conference on Learning Representations (ICLR) (2020).

\bibitem{Jaszczur2021SparseIE} J. Sebastian, C. Aakanksha, M. Afroz, K. Lukasz, G. Wojciech, M. Henryk, and K. Jonni, Sparse is Enough in Scaling Transformers, (2021).

\bibitem{AMRS13} V. Belagiannis, S. Amin, M. AAndriluka, B. Schiele, N. Navab, and S. Ilic, 3D Pictorial Structures Revisited: Multiple Human Pose Estimation, IEEE Transactions on Pattern Analysis and Machine Intelligence, 38 (10) (2016) 1929-1942.

\bibitem{D-AMSB21} A. Diaz-Arias, M. Messmore, D. Shin, and S. Baek, On the role of depth predictions for 3D human pose estimation, https://arxiv.org/abs/2103.02521 (2021).


\bibitem{ConvLSTM} X. Shi, Z. Chen, H. Wang, D. Yeung, W. Wong and W. Woo, Convolutional LSTM Network: A Machine Learning Approach for Precipitation Nowcasting, Proceedings of the 28th International Conference on Neural Information Processing Systems, 1 (2015) 802-810.

\bibitem{Q16} Q. Basatiaan, rnn: a Recurrent Neural Network in R, Working Papers (2016).

\bibitem{hochreiter1997long} S. Hochreiter and H. Schmidhuber, Long short-term memory, Neural computation, 9 (8) (1997) 1735-1780.

\bibitem{li2022mhformer} W. Li, T. Hong, W. Hao, V. Picho, and L. Van Gool, MHFormer: Multi-Hypothesis Transformer for 3D Human Pose Estimation, Proceedings of the IEEE Conference on Computer Vision and Pattern Recognition (CVPR), (2022).

\bibitem{ZT21} Z. Zou and W. Tang, Modulated graph convolutional network for 3D human pose estimation, IEEE International Conference on Computer Vision (ICCV), (2021), 11477-11487.

\bibitem{gong2021poseaug} K. Gong, J. Zhang, and J. Feng, PoseAug: A Differentiable Pose Augmentation Framework for 3D Human Pose Estimation, CVPR, 2021.


\bibitem{li2022exploiting} W. Li, H. Liu, R. Ding, M. Liu, P. Wang, W. Yang Exploiting temporal contexts with strided transformer for 3d human pose estimation, IEEE Transaactions on Multimedia, 2022. 

\bibitem{THUNDR} M. Zanfir, A. Zanfir, E. Bazavan, W. Freeman, R. Sukthankar, and C. Sminchisescu, Thundr: Transformer-based 3d human reconstruction with markers, Proceedings of the IEEE/CVF International Conference on Computer Vision, 2021. 

\bibitem{ptflops} V. Sovrasov, Flops counter for convolutional networks in pytorch framework, https://github.com/sovrasov/flops-counter.pytorch/, 2019-11-12.


\bibitem{MLN19} P. Michel, O. Levy, and G. Neubig, Are Sixteen Heads Really Better than One?, Proceedings of the International Conference on Neural Information Processing Systems (NeurIPS), (2019).

\bibitem{VCCRLB18} Petar Veličković and Guillem Cucurull and Arantxa Casanova and Adriana Romero and Pietro Liò and Yoshua Bengio, Graph Attention Networks, International Conference on Learning Representations (ICLR), 2018 

\bibitem{SDMR20} P. -E. Sarlin, D. DeTone, T. Malisiewicz and A. Rabinovich, "SuperGlue: Learning Feature Matching With Graph Neural Networks," 2020 IEEE/CVF Conference on Computer Vision and Pattern Recognition (CVPR), Seattle, WA, USA, 2020, pp. 4937-4946

\bibitem{Lgraph21} Liu, Junfa and Rojas, Juan and Li, Yihui and Liang, Zhijun and Guan, Yisheng and Xi, Ning and Zhu, Haifei, "A Graph Attention Spatio-temporal Convolutional Network for 3D Human Pose Estimation in Video," 2021 IEEE International Conference on Robotics and Automation (ICRA), Xi'an, China, 2021, pp. 3374-3380

\bibitem{WHXYS20} Jianzhai Wu, Dewen Hu, Fengtao Xiang, Xingsheng Yuan, and Jiongming Su. 2020. 3D human pose estimation by depth map. Vis. Comput. 36, 7 (Jul 2020), 1401–1410

\bibitem{BGK21} S. Banik, A. M. GarcÍa and A. Knoll, "3D Human Pose Regression Using Graph Convolutional Network," 2021 IEEE International Conference on Image Processing (ICIP), Anchorage, AK, USA, 2021, pp. 924-928

\end{thebibliography}
\end{document}